\documentclass[letterpaper]{article} % DO NOT CHANGE THIS

\usepackage{aaai24}  % DO NOT CHANGE THIS
\usepackage{times}  % DO NOT CHANGE THIS
\usepackage{helvet}  % DO NOT CHANGE THIS
\usepackage{courier}  % DO NOT CHANGE THIS
\usepackage[hyphens]{url}  % DO NOT CHANGE THIS
\usepackage{graphicx} % DO NOT CHANGE THIS
\usepackage{natbib}  % DO NOT CHANGE THIS AND DO NOT ADD ANY OPTIONS TO IT
\usepackage{caption} % DO NOT CHANGE THIS AND DO NOT ADD ANY OPTIONS TO IT
\usepackage[utf8]{inputenc} % allow utf-8 input
\usepackage{booktabs}       % professional-quality tables
\usepackage{amsfonts}       % blackboard math symbols
\usepackage{nicefrac}       % compact symbols for 1/2, etc.
\usepackage{microtype}      % microtypography
\usepackage{xcolor}         % colors
\usepackage{multirow}
\usepackage{threeparttable}
\usepackage{amsmath}
\usepackage{amssymb}
\usepackage{bm}
\usepackage{color, colortbl}
\usepackage{subfig}
\usepackage{floatrow}
\usepackage[ruled,vlined]{algorithm2e}
\usepackage{bbding}
\usepackage{dsfont}

 \def\correpond{%
      \ifnum\value{eqfn}=0%
        \footnote{Correspondence to alex.xun.xu@gmail.com and kuijia@cuhk.edu.cn.}%
        \setcounter{eqfn}{\value{footnote}}%
      \else%
        \footnotemark[\value{eqfn}]%
      \fi%
    }%

\newfloatcommand{capbtabbox}{table}[][\FBwidth]

\newcommand{\method}{\textbf{TRIBE}}

\newcommand{\vect}[1]{\mathbf{#1}}
\newcommand{\matr}[1]{\mathbf{#1}}

\newcommand{\change}[1]{\textcolor{black}{#1}}

\definecolor{Gray}{gray}{0.9}

\definecolor{text01purple}{RGB}{168,119,200}

\definecolor{text01green}{RGB}{82,208,83}
\newcommand{\textgreen}[1]{\textcolor{text01green}{#1}}

\definecolor{text02red}{RGB}{211,41,15}

\definecolor{text02yellow}{RGB}{230,119,11}

\newcommand{\dtplus}[1]{\fontsize{6pt}{0.1em}\selectfont (\textbf{\textgreen{#1}})\fontsize{10pt}{1em}}

\usepackage{newfloat}
\usepackage{listings}
\DeclareCaptionStyle{ruled}{labelfont=normalfont,labelsep=colon,strut=off} % DO NOT CHANGE THIS
\floatstyle{ruled}
\newfloat{listing}{tb}{lst}{}
\floatname{listing}{Listing}
\title{Towards Real-World Test-Time Adaptation: Tri-net Self-Training with Balanced Normalization}
\author{
    Yongyi Su\textsuperscript{\rm 1},
    Xun Xu\textsuperscript{\rm 2, 1}\correpond,
    Kui Jia\textsuperscript{\rm 3}\correpond
}
\affiliations{
    \textsuperscript{\rm 1}South China University of Technology\\
    \textsuperscript{\rm 2}Institute for Infocomm Research, A*STAR\\
    \textsuperscript{\rm 3}School of Data Science, The Chinese University of Hong Kong, Shenzhen \\
    eesuyongyi@mail.scut.edu.cn, alex.xun.xu@gmail.com, kuijia@cuhk.edu.cn
}

\begin{document}

\maketitle

\begin{abstract}
Test-Time Adaptation aims to adapt source domain model to testing data at inference stage with success demonstrated in adapting to unseen corruptions. However, these attempts may fail under more challenging real-world scenarios. Existing works mainly consider real-world test-time adaptation under non-i.i.d. data stream and continual domain shift. In this work, we first complement the existing real-world TTA protocol with a globally class imbalanced testing set. We demonstrate that combining all settings together poses new challenges to existing methods. We argue the failure of state-of-the-art methods is first caused by indiscriminately adapting normalization layers to imbalanced testing data. To remedy this shortcoming, we propose a balanced batchnorm layer to swap out the regular batchnorm at inference stage. The new batchnorm layer is capable of adapting without biasing towards majority classes. We are further inspired by the success of self-training~(ST) in learning from unlabeled data and adapt ST for test-time adaptation. However, ST alone is prone to over adaption which is responsible for the poor performance under continual domain shift. Hence, we propose to improve self-training under continual domain shift by regularizing model updates with an anchored loss. The final TTA model, termed as TRIBE, is built upon a tri-net architecture with balanced batchnorm layers. We evaluate TRIBE on four datasets representing real-world TTA settings. TRIBE consistently achieves the state-of-the-art performance across multiple evaluation protocols. 
The code is available at \url{https://github.com/Gorilla-Lab-SCUT/TRIBE}. 

\end{abstract}

% \vspace{-0.4cm}

\section{Introduction}

\begin{figure*}[!htb]
    \centering
    \includegraphics[width=0.95\linewidth]{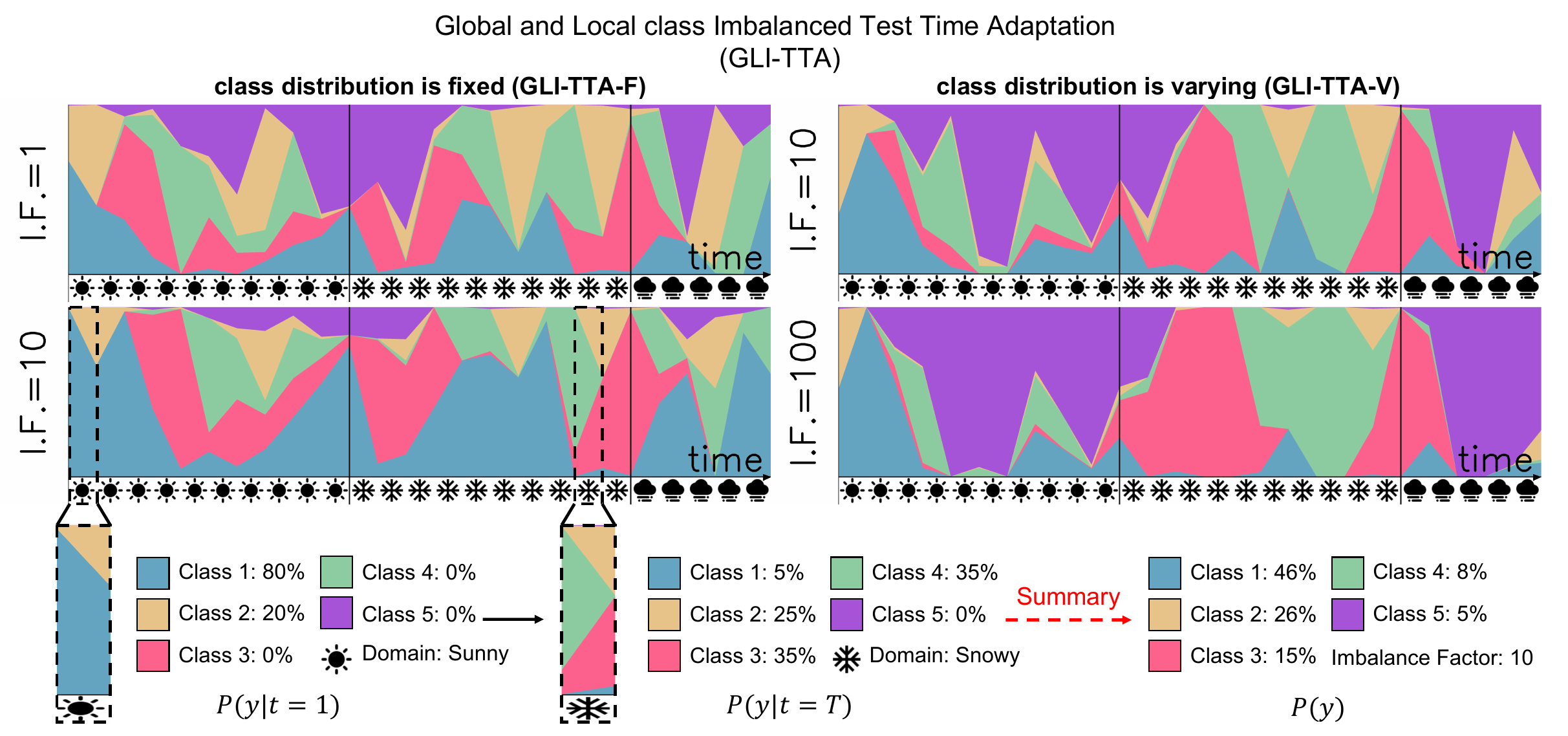}
    \caption{{{Illustration of two challenging real-world TTA scenarios. Different colors indicate the proportions of semantic classes, horizontal axis indicates testing data domain~(e.g. different corruptions) may shift over time and different imbalance factor~($I.F.$) controls the degree of global imbalance. We expect the testing data stream to exhibit both local and global class imbalance, termed as ``class distribution is fixed~(GLI-TTA-F)'' and this distribution may also evolve over time, termed as ``class distribution is varying~(GLI-TTA-V)''.}}
    % \vspace{-0.6cm}
    } 
    % \vspace{-0.4cm}
    \label{fig:Setting}
\end{figure*}

The recent success of deep neural networks relies on the assumption of generalizing pre-trained model to i.i.d. testing domain~\cite{wang2022generalizing}. When deep learning models are to be deployed on real-world applications, robustness to out-of-distribution testing data, e.g. visual corruptions caused by lighting conditions, adverse weather, etc. becomes a major concern. Recent studies revealed such corruptions could severely deteriorate the generalization of model pre-trained on clean training samples~\cite{ModelNet40-C,hendrycks2018benchmarking,sakaridis2018semantic}. Importantly, the \change{corruption} on testing data is often unknown and sometimes unpredictable before deployment. Therefore, a new line of works emerge by adapting pre-trained models to testing data distribution at inference stage, a.k.a. test-time adaptation~(TTA)
\cite{sun2020test,tent_wang2020,su2022revisiting}. The success of test-time adaptation is often achieved by distribution alignment~\cite{su2022revisiting,liu2021ttt++}, self-supervised training~\cite{chen2022contrastive} and self-training~\cite{goyaltest2022}, all demonstrating remarkable improvement of robustness on multiple types of visual corruptions in the testing data. Despite the unprecedented performance, existing TTA approaches are often developed under restrictive assumptions of testing data, e.g. stationary class distribution and static domain shift, and this gives rise to many attempts to explore TTA methods for real-world testing data ~\cite{cotta,yuan2023robust,note,niu2023towards}.

The recently explored real-world TTA, a.k.a. wild TTA~\cite{niu2023towards} or Practical TTA~\cite{yuan2023robust}, settings mainly consider the challenges brought by local class-imbalance~\cite{niu2023towards,yuan2023robust,note} and continual domain shift~\cite{cotta} which are expected to be encountered in real-world applications. %and/or update with small batchsize~\cite{niu2023towards}. 
Local class-imbalance is often observed when testing data are drawn in a non-i.i.d. manner~\cite{note}. Direct adaptation indiscriminately results in biased distribution estimation and the recent works proposed exponential batchnorm update~\cite{yuan2023robust} or instance batchnorm update~\cite{note} to tackle this challenge.
In this work, \change{our aim is} to address beyond the local class-imbalance challenge by taking into account the fact that the global distribution of testing data could be severely imbalanced and the class distribution may shift over time. We provide an illustration of the more challenging scenario in Fig.~\ref{fig:Setting}. This additional challenge renders existing TTA methods ineffective as the class prevalence on testing data is unknown before inference stage and the model could be biased towards majority classes through blind test-time adaptation. Through empirical observations, this issue becomes particularly acute for methods relying on estimating global statistics for updating normalization layers\cite{BN_Stat,PL,tent_wang2020}. It mainly owes to the fact that a single global distribution is estimated from the whole testing data on which samples are normalized. As such, the global distribution could easily bias towards majority classes, resulting in internal covariate shift~\cite{ioffe2015batch}. To avoid biased batch normalization~(BN), we propose a balanced batch normalization layer by modeling the distribution for each individual category and the global distribution is extracted from category-wise distributions. The balanced BN allows invariant estimation of distribution under both locally and globally class-imbalanced testing data.

Shift of domain over time occurs frequently in real-world testing data, e.g. a gradual change of lighting/weather conditions. It poses another challenge to existing TTA methods as the model could overly adapt to domain A and struggle with domain B when A shifts to B. To alleviate overly adapting to a certain domain, CoTTA~\cite{cotta} randomly reverts model weights to pre-trained weights and EATA~\cite{niu2022efficient} regularizes the adapted model weights against source pre-trained weights to avoid overly shifting model weights. Nevertheless, these approaches still do \change{not} explicitly address the challenge of constant shifting domains in testing data. 
As self-training has been demonstrated to be effective for learning from unlabeled data~\cite{sohn2020fixmatch}, we adopt a teacher-student framework for TTA. Nonetheless, direct self-training without regularization is prone to confirmation bias~\cite{arazo2020pseudo} and could easily overly adapt pre-trained model to a certain domain, causing degenerate performance upon seeing new domains. To avoid this over adaptation, we further introduce an anchor network, of which the weights are copied from pre-trained model and batchnorm layers are dynamically updated by testing samples. The anchored loss, realised as mean square error~(MSE), between teacher and anchor network is jointly optimised with self-training loss to strike a balance between adaptation to specific domain and being versatile on ever changing domains. We brand this design as a tri-net architecture. 
We demonstrate that with the help of tri-net, TTA maintains a good performance within a wider range of learning rate. We refer to the final model as \textbf{TRI}-net self training with \textbf{B}alanc\textbf{E}d normalization~(\textbf{TRIBE}) in recognition of the tri-net architecture with balanced normalization layer.

We summarize the contributions of this work as follows.
    % \vspace{-0.1cm}

\begin{itemize}
        % \vspace{-0.1cm}
    \item We are motivated by the challenges in real-world test-time adaptation and propose to tackle a challenging TTA setting where testing data is both locally and globally class-imbalanced and testing domain may shift over time. 
        % \vspace{-0.1cm}
    \item A novel balanced batch normalization layer is introduced to fit to testing data distribution with both local and global class imbalance.
        % \vspace{-0.1cm}
    \item We further introduce a tri-net framework to facilitate adaptation under continually shifting testing domain. We demonstrate this tri-net design improves robustness to the choice of learning rate.
        % \vspace{-0.1cm}
    \item We evaluate the proposed method, TRIBE, on four test-time adaptation datasets under different real-world scenarios, demonstrating superior performance to all state-of-the-art methods.
            % \vspace{-0.1cm}
\end{itemize}

\begin{figure*}[!t]
    \centering
    \includegraphics[width=0.95\linewidth]{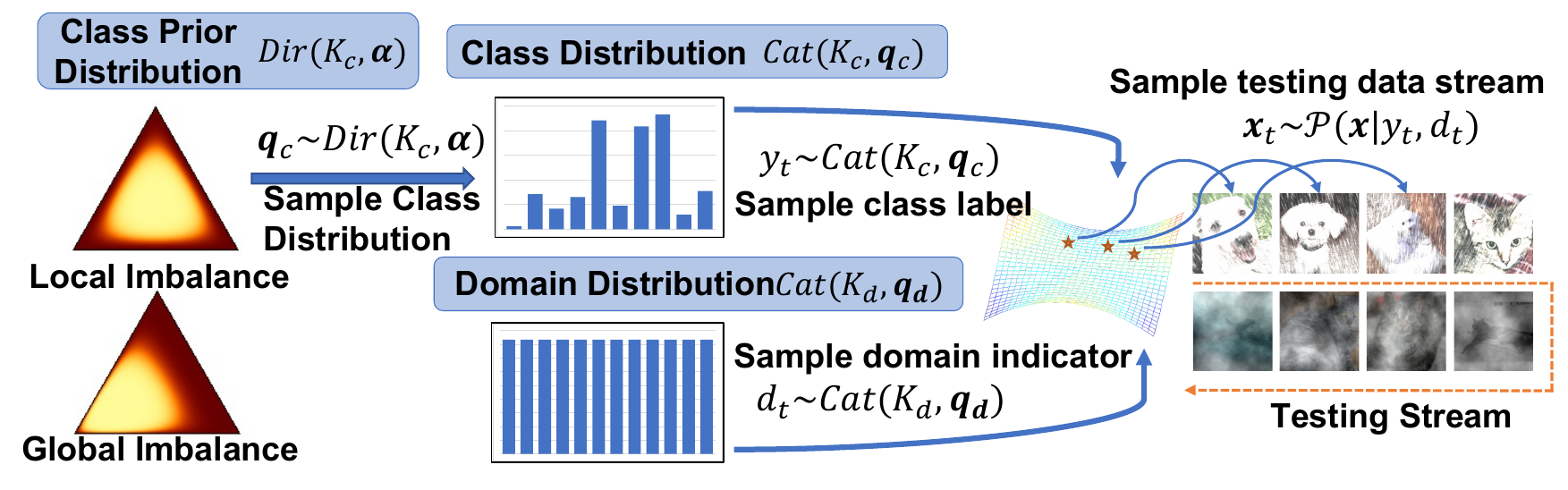}
    \caption{{An illustration of the proposed real-world TTA simulation protocol with a hierarchical probabilistic model. A non-uniform $\alpha$ results in globally imbalanced testing data distribution.}
    % \vspace{-0.7cm}
    }
    % \vspace{-0.3cm}
    \label{fig:protocol}    
\end{figure*}

% \vspace{-0.4cm}
\section{Related Work}
\noindent\textbf{Unsupervised Domain Adaptation}:
Machine learning models often assume both training and testing data are drawn i.i.d. from the same distribution. When such assumption is violated, generalizing source model to testing distribution is hampered by the domain shift, leading to degraded performance~\cite{wang2018deep}. Unsupervised domain adaptation~(UDA) improves model generalization by exploiting both labeled source domain data and unlabeled target domain data~\cite{ganin2015unsupervised,tzeng2014deep,long2015learning}. Common approaches towards UDA includes distribution alignment~\cite{gretton2012kernel,sun2016deep,zellinger2017central}, adversarial learning~\cite{hoffman2018cycada}, target clustering~\cite{tang2020unsupervised} and self-training~\cite{liu2021cycle}. Nevertheless, UDA is only effective when source and target domain data are simultaneously accessible. More importantly, in real-world applications the distribution in target domain is often predictable until inference stage which has motivated research into test-time adaptation.

\noindent\textbf{Test-Time Adaptation}:
Adapting pre-trained model to target domain distribution at test-time improves model generalization to unseen distribution shift. Widely adopted test-time adaptation~(TTA) protocol simultaneously evaluate on a stream of testing data and update model weights~\cite{sun2020test,tent_wang2020,iwasawa2021test,su2022revisiting,gandelsman2022test,goyaltest2022,chen2022contrastive}. The state-of-the-art approaches towards TTA adopt self-training~\cite{tent_wang2020,su2023revisiting,gandelsman2022test}, distribution alignment~\cite{sun2020test,su2022revisiting} and self-supervised learning~\cite{liu2021ttt++,chen2022contrastive}. With the above techniques, generalization performance on testing data with corruptions has been substantially improved. Nonetheless, most of these are optimized towards the vanilla TTA protocol, thus these methods may not maintain the superior performance under more realistic TTA scenarios.

\noindent\textbf{Real-World Test-Time Adaptation}:
Deploying TTA methods in real-world application requires tackling commonly encountered challenges. Recent works summarized multiple challenges that could appear in real-world test-time adaptation, including updating with small batchsize~\cite{niu2023towards}, non-i.i.d. or temporally correlated testing data~\cite{note,cotta,yuan2023robust,niid_boudiaf2022parameter} and continually adapting to shifting domains~\change{\cite{cotta,yuan2023robust,brahma2023probabilistic}}. Empirical observations \change{demonstrate} that these real-world challenges could pose great challenges to existing TTA methods. Despite the recent efforts in developing TTA robust to non-i.i.d. testing data, we argue that a systematic investigation into more diverse real-world challenges, including global class-imbalance, is missing. This work propose a principled way to simulate these challenges and develop a self-training based method with balanced batchnorm to achieve the state-of-the-art performance.

% \vspace{-0.2cm}
\section{Methodology}

\subsection{Real-World TTA Protocol}

We denote a stream of testing data as $\matr{x}_0,\matr{x}_1,\cdots,\vect{x}_T$ where each $\vect{x}_t$ is assumed to be drawn from a distribution $\mathcal{P}(\vect{x}|d_t,y_t)$ conditioned on two time-varying variables, namely the testing domain indicator $d_t\in\{1,\cdots K_d\}$ and the class label $y_t\in\{1,\cdots K_c\}$, where $K_d$ and $K_c$ refer to the number of domains (e.g. type of corruptions) and number of semantic classes.  
In the real-world TTA protocol, both the testing domain indicator and class label distribution could be subject to constant shift, in particular, we assume the domain indicator to exhibit a gradual and slowly shift over time. This is manifested by many real-world applications, e.g. the lighting and weather conditions often changes slowly. We further point out that testing samples are often class imbalanced both locally within a short period of time and globally over the whole testing data stream. Therefore, we model the testing data stream as sampling from a hierarchical probabilistic model. Specifically, we denote a prior $\vect{\alpha}\in\mathds{R}^{K_c}$ parameterizing a Dirichlet distribution $\vect{q}_c\sim Dir(K_c,\vect{\alpha})$. Within a stationary local time window, e.g. a minibatch of testing samples, the labels of testing samples are drawn from a categorical distribution $y\sim Cat(K_c,\vect{q}_c)$ where $\vect{q}_c$ is drawn from the conjugate prior distribution  $Dir(K_c,\vect{\alpha})$. The corrupted testing sample is then assumed to be finally sampled from a complex distribution conditioned on the domain indicator $d_t$ and class label $y_t$, written as $\vect{x}\sim\mathcal{P}(\vect{x}|d_t,y_t)$. The domain indicator can be modeled as another categorical distribution parameterized by a fixed probability $d_t\sim Cat(K_d,\vect{q}_d)$. A hierarchical probabilistic model simulating the real-world TTA protocol is presented in Fig.~\ref{fig:protocol}. We notice the probabilistic model can instantiate multiple existing TTA protocols. For instance, when testing data are locally class imbalanced, as specified by~\cite{yuan2023robust,note}, a uniform proportion parameter $\alpha=\sigma\mathbf{1}$ is chosen with a scale parameter $\sigma$ controlling the degree of local imbalanceness and $\vect{q}_c$ is re-sampled every mini-batch. We can easily simulate global class-imbalance by specifying a non-uniform $\alpha$.  
We defer a more detailed discussion of simulating real-world TTA protocols with the hierarchical probabilistic model to the supplementary.

\begin{figure*}[t]
    \centering
    \includegraphics[width=0.95\linewidth]{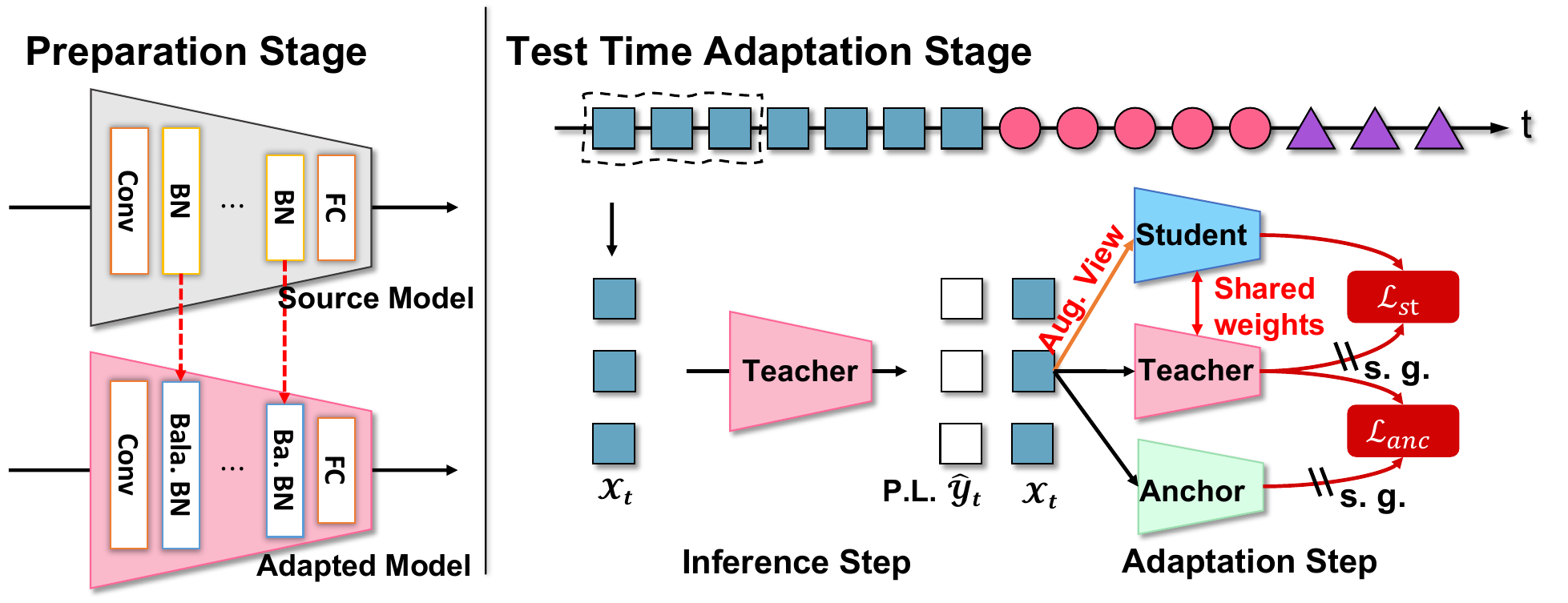}
    \caption{{Illustration of the proposed method. We replace the Batchnorm layer of the source model with our proposed Balanced Batchnorm for imbalanced testing set. During test time adaptation, we optimize the combination of self-training loss $\mathcal{L}_{st}$ and anchor loss $\mathcal{L}_{anc}$.}
    % \vspace{-0.5cm}
    } 
    % \vspace{-0.3cm}
    \label{fig:Method}
\end{figure*}

% \vspace{-0.3cm}
\subsection{Balanced Batch Normalization}

Batch normalization~(BN)~\cite{ioffe2015batch} plays a critical role in enabling more stable model training, reducing sensitivity to hyper-parameters and initialization. When testing data features a distribution shift from the source training data, the regular practice of freezing BN statistics for inference fails to maintain the generalization~\cite{yuan2023robust, BN_Stat, note, lim2023ttn, niu2023towards}. Hence, adapting BN statistics to testing data distribution becomes a viable solution to TTA~\cite{BN_Stat}. To adapt model subject to locally imbalanced testing data, the robust batch normalization~\cite{yuan2023robust} updates BN's mean and variance in a moving average manner on testing data. The robust BN smooths out the bias estimated within each minibatch and achieves competitive performance under non-i.i.d. testing data stream. 

Despite being successful in non-i.i.d. testing data stream, a naive moving average update policy struggles in adapting to globally imbalanced testing domain. For example, evidenced in the empirical evaluation in Tab.~\ref{tab:global_imbalance_cifar10}, the performance of RoTTA~\cite{yuan2023robust} degenerates substantially under more severely global imbalanced testing data. We ascribe the poor performance to the fact that a single BN will bias towards majority classes and normalizing samples from the minority classes with biased statistics will result in severe covariate shift within internal representations. This will eventually cause mis-classifying the minority classes and lower the macro average accuracy. To remedy the bias in adapting BN statistics, we propose a Balanced Batchnorm layer which maintains $K_c$ pairs of statistics separately for each semantic class, denoted as $\{\mu_k\}_{k=1\cdots K_c}$, $\{\sigma_k\}_{k=1\cdots K_c}$. To update category-wise statistics, we apply an efficient iterative updating approach with the help of pseudo labels predictions as follows,
    % \vspace{-0.3cm}

\begin{equation}
\resizebox{0.9\linewidth}{!}{
$
\begin{split}
    &\mu_k^t = \mu_k^{t-1} + \delta_k, \\
    &{\sigma^2_k}^t = {\sigma^2_k}^{t-1} - \delta_k^2 + \eta \sum_{b=1}^B {\mathds{1}(\hat{y}_b = k) \frac{1}{HW}\sum_{h=1}^H \sum_{w=1}^W \left[ ({F}_{bhw} - \mu_k^{t-1})^2 - {\sigma^2_k}^{t-1} \right]}\\
    &s.t.\quad \delta_k = \eta \sum_{b=1}^B {\mathds{1}(\hat{y}_b = k)\frac{1}{HW} \sum_{h=1}^H\sum_{w=1}^W({F}_{bhw} - \mu_k^{t-1})},
\end{split}
$
}
\end{equation}
where $\matr{F}\in \mathds{R}^{B\times C\times H\times W}$ denotes the input for Balanced BN layer and $\hat{y}_b$ is the pseudo label predicted by the adapted model {in the inference step}. With the above design BN statistics for each individual class is separately updated and the global BN statistics are derived from all category-wise statistics as in Eq.~\ref{eq:bn_global}.
    % \vspace{-0.3cm}

\begin{equation}\label{eq:bn_global}
\resizebox{0.9\linewidth}{!}{
$
% \begin{split}
    \mu_g = \frac{1}{K_c} \sum_{k=1}^{K_c} \mu_k^t, \quad \sigma_g^2 = \frac{1}{K_c} \sum_{k=1}^{K_c} \left[{\sigma_k^2}^t + (\mu_g - \mu_k^t)^2\right].
% \end{split}
$
}
\end{equation}

Nevertheless, we found when the number of categories is large or the pseudo labels are highly untrustworthy, e.g. the baseline accuracy on ImageNet-C is very low, the above updating strategy might be less effective due to its reliance on the pseudo labels. Therefore, we combine the class-agnostic updating strategy (Robust BN) and the category-wise updating strategy with a balancing parameter $\gamma$ as below.
    % \vspace{-0.3cm}

\begin{equation}
\resizebox{0.8\linewidth}{!}{
$
    \mu_k^t = \mu_k^{t-1} + (1-\gamma) \delta_k + \gamma \frac{1}{K_c} \sum_{k'=1}^{K_c} \delta_{k'},
$
}
\end{equation}
% \vspace{-0.2cm}
\begin{equation}
\resizebox{0.9\linewidth}{!}{
$
\begin{split}
  & {\sigma^2_k}^t = {\sigma^2_k}^{t-1} + \\
  & (1-\gamma) \left\{-\delta_k^2 + \eta \sum_{b=1}^B {\mathds{1}(\hat{y}_b = k)\frac{1}{HW} \sum_{h=1}^H\sum_{w=1}^W \left[ ({F}_{bhw} - \mu_k^{t-1})^2 - {\sigma^2_k}^{t-1} \right]}\right\} + \\
     & \gamma \cdot \frac{1}{K_c} \sum_{k'=1}^{K_c} \left\{-\delta_{k'}^2 + \eta\sum_{b=1}^B {\mathds{1}(\hat{y}_b = k') \frac{1}{HW} \sum_{h=1}^H\sum_{w=1}^W\left[ ({F}_{bhw} - \mu_{k'}^{t-1})^2 - {\sigma^2_k}^{t-1} \right]}\right\}.
\end{split}
$
}
\end{equation}

Specifically, when $\gamma=0$ the updating strategy is the pure class-wise updating strategy and when $\gamma=1$ the updating strategy degrades to the rule in Robust BN. In all experiments of this paper, we leverage $\gamma=0.0$ in CIFAR10-C, $\gamma=0.1$ in CIFAR100-C due to the large number of class and $\gamma=0.5$ in ImageNet-C due to the highly untrustworthy pseudo labels. The instance-level momentum coefficient $\eta$ in Balanced BN is set to $0.0005 \times K_c$.

% \vspace{-0.2cm}
\subsection{Tri-Net Self-Training}

Self-Training (ST) has demonstrated tremendous effectiveness in multiple tasks~\cite{sohn2020fixmatch, kumar2020understanding}. ST updates the model through constraining the prediction consistency between original samples and corresponding augmented samples. 
In this work, we adopt an approach similar to semi-supervised learning~\cite{sohn2020fixmatch} to fine-tune the model to adapt the testing data. In specific, as illustrated in Fig.~\ref{fig:Method}, we introduce teacher $f_t(\vect{x};\Theta)$ and student $f_s(\vect{x};\Theta)$ networks where the BN layers are independently updated while other weights are shared. The pseudo labels for testing sample are predicted by the teacher network and only the confident pseudo labels are employed for training the student network. Specifically, we denote the probabilistic posterior as $\vect{p}=h(f(\vect{x}))$ and define the self-training loss in Eq.~\ref{eq:selftrain_loss}, where $\vect{p}^s=h(f_s(\mathcal{A}(\vect{x}); \Theta)), \vect{p}^t=h(f_t(\vect{x}; \Theta))$, $\hat{\vect{p}}^t$ refers to the one-hot pseudo label of $\vect{p}^t$, $\mathcal{A}$ refers to a strong data augmentation operation, $\mathcal{H}$ refers to entropy and cross-entropy losses and $H_0$ defines a thresholding hyper-parameter.
    % \vspace{-0.3cm}

\begin{equation}\label{eq:selftrain_loss}
    \mathcal{L}_{st} = \frac{\sum_{b=1}^B \mathds{1}(\mathcal{H}(\matr{p}^{t}_b) < H_0 \cdot \log K_c) \cdot \mathcal{H}( \hat{\vect{p}}^{t}_b, \vect{p}^s_b)}{\sum_{b=1}^B \mathds{1}(\mathcal{H}(\matr{p}^{t}_b) < H_0 \cdot \log K_c)}
\end{equation}

A recent study revealed that self-training is effective for TTA~\cite{su2023revisiting}, however without additional regularizations self-training is easily subject to confirmation bias~\cite{arazo2020pseudo}. This issue would only exacerbate when test data distribution is highly imbalanced, thus leading to over adaptation or collapsed predictions. To avoid over adapting model to a certain test domain, we further propose to incorporate an additional network branch as anchor for regularization.

\noindent\textbf{Anchor Network}: We use a frozen source domain network as the anchor network to regularize self-training. In particular, we copy the source model weights, freeze all weights and swap regular BN layers with the proposed Balanced BN layers. To regularize self-training, we design an anchored loss as the mean square error between the posterior predictions of teacher and anchor networks as in Eq.~\ref{eq:anchorloss}.
As three network branches are jointly utilized, it gives rise to the term of tri-net self-training.
% \vspace{-0.2cm}
\begin{equation}\label{eq:anchorloss}
    \mathcal{L}_{anc} = \frac{\sum_{b=1}^B\mathds{1}(\mathcal{H}(\matr{p}^{t}_b) < H_0 \cdot \log K_c) ||\vect{p}^t_b - \vect{p}^a_b||_2^2 }{K_c \sum_{b=1}^B\mathds{1}(\mathcal{H}(\matr{p}^{t}_b) < H_0 \cdot \log K_c)}
\end{equation}

We finally simultaneously optimize the self-training and anchored losses $\mathcal{L}=\mathcal{L}_{st}+\lambda_{anc}\mathcal{L}_{anc}$ w.r.t. the affine parameters of the Balanced BN layers for efficient test-time adaptation.
% \vspace{-0.4cm}
% \vspace{-0.3cm}
\section{Experiment}

\subsection{Experiment Settings}

\noindent\textbf{Datasets}:
We evaluate on four test-time adaptation datasets, including \textbf{CIFAR10-C}~\cite{hendrycks2018benchmarking}, \textbf{CIFAR100-C}~\cite{hendrycks2018benchmarking}, \textbf{ImageNet-C}~\cite{hendrycks2018benchmarking} and \textbf{MNIST-C}~\cite{mu2019mnist}. Each of these benchmarks includes 15 types of corruptions with 5 different levels of severity.  
CIFAR10/100-C both have 10,000 testing samples evenly divided into 10/100 classes for each type of corruptions. ImageNet-C has 5,000 testing samples for each corruption unevenly divided into 1,000 classes~\footnote{ImageNet-C is only evaluated in a subset with 5,000 testing samples on RobustBench: https://github.com/RobustBench/robustbench}. We evaluate all methods under the largest corruption severity level 5 and report the classification error rate ($\%$) throughout the experiment section. We include the detailed results of \textbf{MNIST-C}~\cite{mu2019mnist} in the supplementary.

\noindent\textbf{Hyper-parameters}: For CIFAR10-C and CIFAR100-C experiments, we follow the official implementations from previous TTA works~\cite{tent_wang2020, cotta, yuan2023robust} and respectively adopt a standard pre-trained WideResNet-28~\cite{zagoruyko2016wide} and ResNeXt-29~\cite{xie2017aggregated} models from RobustBench~\cite{croce2robustbench} benchmark, for the fair comparison.  
For ImageNet-C experiments, the standard pre-trained ResNet-50~\cite{he2016deep} model in torchvision is adopted. For most competing methods and our TRIBE, we leverage the Adam~\cite{kingma2014adam} optimizer with the learning rate 1e-3 in CIFAR10/100-C and ImageNet-C experiments. As an exception, for Note~\cite{note} and TTAC~\cite{su2022revisiting} we use the learning rate released in their official implementations. We use a batchsize of 64 for CIFAR10/100-C and 48 for ImageNet-C. Other hyper-parameters of our proposed model are listed as follow: $\lambda_{anc}=0.5, \eta=0.0005 \times {K_c}$ in all datasets, in CIFAR10-C $H_0=0.05, \gamma=0.$, in CIFAR100-C $H_0=0.2, \gamma=0.1$ and in ImageNet-C $H_0=0.4, \gamma=0.5$. Adequate hyper-paramter analysis, provided in the supplementary, demonstrate that the hyper-parameters used into TRIBE are not sensitive. The data augmentations used in TRIBE are described in the supplementary. All of our experiments can be performed on a single NVIDIA GeForce RTX 3090 card.

\noindent\textbf{TTA Evaluation Protocol}: We evaluate under two real-world TTA protocols, namely the \textbf{GLI-TTA-F} and \textbf{GLI-TTA-V}. For both protocols, we create a global class imbalanced testing set following the long-tail dataset creation protocol~\cite{cui2019class}, we choose three imbalance factor $I.F.$ as \textcolor{black}{1, 10, 100 and 200} for evaluation \textcolor{black}{where GLI-TTA degrades into PTTA setting~\cite{yuan2023robust} with $I.F.=1$ }. A default scale parameter $\sigma=0.1$ is chosen to control local class imbalance. To simulate continually shifting domains, we sample without replacement the domain indicator after all testing samples are predicted. For better reproducibility we provide the sequence of domains in the supplementary. Under the GLI-TTA-F setting, we fix the proportion parameter $\alpha$ throughout the experiment. Under the GLI-TTA-V setting, we randomly permute class indices after adaptation to each domain (type of corruption) to simulate time-varying class distribution.

\begin{table}[t]
    \centering
    \caption{{Average classification error on CIFAR10-C while continually adapting to different corruptions at the highest severity 5 with globally and locally class-imbalanced test stream. $I.F.$ is the Imbalance Factor of Global Class Imbalance. Instance-wise average error rate $a\%$ and  category-wise average error rate $b\%$ are separated by (a / b).}}
    \resizebox{\linewidth}{!}{
        \begin{tabular}{l|c|c|c|c}
        \toprule[1.2pt]
        \multirow{2}{*}{Method} & \multicolumn{4}{c}{\textbf{Fixed Global Class Distribution (GLI-TTA-F)}}\\
        & $I.F.=1$ & $I.F.=10$ & $I.F.=100$ & $I.F.=200$ \\
        \midrule
        TEST  & 43.50 / 43.50 & 42.64 / 43.79 & 41.71 / 43.63 & 41.69 / 43.47 \\
        BN
        % ~\cite{BN_Stat}
        & 75.20 / 75.20 & 70.77 / 66.77 & 70.00 / 50.72 & 70.13 / 47.34\\
        PL
        % ~\cite{PL}
        & 82.90 / 82.90 & 72.43 / 70.59 & 70.09 / 55.29 & 70.38 / 49.86 \\
        % TENT~(ICLR21)
        TENT
        % ~\cite{tent_wang2020}
        & 86.00 / 86.00 & 78.15 / 74.90 & 71.10 / 58.59 & 69.15 / 53.37 \\
        % LAME~(CVPR22)
        LAME
        % ~\cite{niid_boudiaf2022parameter}
        & 39.50 / 39.50 & 38.45 / 40.07 & 37.48 / 41.80 & 37.52 / 42.59 \\
        % \change{COTTA}~(CVPR22)
        COTTA
        % ~\cite{cotta}
        & \change{83.20 / 83.20} & \change{73.64 / 71.48} & \change{71.32 / 56.44} & \change{70.78 / 49.98}\\ 
        % NOTE~(NIPS22)
        NOTE
        % ~\cite{note}
        & 31.10 / 31.10 & 36.79 / 30.22 & 42.59 / 30.75 & 45.45 / 31.17 \\
        % TTAC~(NIPS22)
        TTAC
        % ~\cite{su2022revisiting}
        & 23.01 / 23.01 & 31.20 / 29.11 & 43.40 / 37.37 & 46.27 / 38.75 \\
        % \change{PETAL}~(CVPR23)
        PETAL
        % ~\cite{brahma2023probabilistic}
        & \change{81.05 / 81.05} & \change{73.97 / 71.64} & \change{71.14 / 56.11} & \change{71.05 / 50.57}\\
        % RoTTA~(CVPR23)
        RoTTA
        % ~\cite{yuan2023robust}
        & 25.20 / 25.20 & 27.41 / 26.31 & 30.50 / 29.08 & 32.45 / 30.04\\
        \midrule
        % \method & \bf 16.14\dtplus{+6.86} / \bf 16.14\dtplus{+6.86} & \bf 20.98\dtplus{+6.43} / \bf 22.49\dtplus{+3.82} & \bf 19.53\dtplus{+10.97} / \bf 24.66\dtplus{+4.42} & \bf 19.16\dtplus{+13.29} / \bf 24.00\dtplus{+6.04} \\
        \method & \bf 16.14 / \bf 16.14 & \bf 20.98 / \bf 22.49 & \bf 19.53 / \bf 24.66 & \bf 19.16 / \bf 24.00 \\

        \midrule[1.2pt]
        \midrule[1.2pt]

        \multirow{2}{*}{Method} & \multicolumn{4}{c}{\textbf{Time-Varying Global Class Distribution (GLI-TTA-V)}}\\
        & $I.F.=1$ & $I.F.=10$ & $I.F.=100$ & $I.F.=200$ \\
        \midrule

        TEST  & 43.50 / 43.50 &  41.95 / 43.65 & 40.74 / 43.83 & 40.53 / 43.77 \\
        BN
        % ~\cite{BN_Stat}
        & 75.20 / 75.20 & 71.36 / 67.70 & 70.35 / 53.07 & 70.88 / 50.67 \\
        PL
        % ~\cite{PL}
        & 82.90 / 82.90 & 74.74 / 72.12 & 73.03 / 57.53 & 72.49 / 54.20 \\
        % TENT~(ICLR21)
        TENT
        % ~\cite{tent_wang2020}
        & 86.00 / 86.00 & 77.69 / 74.23 & 72.99 / 58.65 & 73.45 / 54.96 \\
        % LAME~(CVPR22)
        LAME
        % ~\cite{niid_boudiaf2022parameter}
        & 39.50 / 39.50 & 38.02 / 40.15 & 36.51 / 42.16 & 36.24 / 42.16 \\
        % \change{COTTA}~(CVPR22)
        COTTA
        % ~\cite{cotta}
        & \change{83.20 / 83.20} & \change{75.29 / 71.87} & \change{73.83 / 56.80} & \change{74.97 / 56.47}\\ 
        % NOTE~(NIPS22)
        NOTE
        % ~\cite{note}
        & 31.10 / 31.10 & 29.52 / 29.23 & 30.02 / 29.88 & 29.71 / 30.28 \\
        % TTAC~(NIPS22)
        TTAC
        % ~\cite{su2022revisiting}
        & 23.01 / 23.01 & 32.25 / 32.12 & 36.84 / 37.13 & 37.96 / 38.07 \\
        % \change{PETAL}~(CVPR23)
        PETAL
        % ~\cite{brahma2023probabilistic}
        & \change{81.05 / 81.05} & \change{75.19 / 71.65} & \change{72.71 / 55.73} & \change{73.76 / 53.51}\\
        % RoTTA~(CVPR23)
        RoTTA
        % ~\cite{yuan2023robust}
        & 25.20 / 25.20 & 27.61 / 26.35 & 32.16 / 29.32 & 33.34 / 31.35 \\
        \midrule
        % \method & \bf 16.14\dtplus{+6.86} / \bf 16.14\dtplus{+6.86} & \bf 20.92\dtplus{+6.69} / \bf 22.40\dtplus{+3.95} & \bf 22.44\dtplus{+7.58} / \bf 25.50\dtplus{+3.82} & \bf 23.10\dtplus{+6.61} / \bf 27.03\dtplus{+3.25} \\
        \method & \bf 16.14 / \bf 16.14 & \bf 20.92 / \bf 22.40 & \bf 22.44 / \bf 25.50 & \bf 23.10 / \bf 27.03 \\
        
        \bottomrule[1.2pt]
        \end{tabular}
    }
    % \vspace{-0.3cm}
    \label{tab:global_imbalance_cifar10}
\end{table}

\begin{table}[!]
    \centering
    \caption{{Average classification error on CIFAR100-C while continually adapting to different corruptions at the highest severity 5 with globally and locally class-imbalanced test stream.}
    % \vspace{-0.7cm}
    }
    % \vspace{-0.5cm}
    \resizebox{\linewidth}{!}{
        \begin{tabular}{l|c|c|c|c}
        \toprule[1.2pt]
        \multirow{2}{*}{Method} & \multicolumn{4}{c}{\textbf{Fixed Global
        Class Distribution (GLI-TTA-F)}}\\
        & $I.F.=1$ & $I.F.=10$ & $I.F.=100$ & $I.F.=200$ \\
        \midrule
        TEST  & 46.40 / 46.40 & 46.96 / 46.52 & 47.53 / 45.91 & 47.59 / 39.94 \\
        BN
        % ~\cite{BN_Stat}
        & 52.90 / 52.90 & 46.05 / 42.29 & 47.01 / 40.01 & 47.38 / 35.26 \\
        PL
        % ~\cite{PL}
        & 88.90 / 88.90 & 68.51 / 69.71 & 53.46 / 57.26 & 49.41 / 49.26 \\
        % TENT~(ICLR21)
        TENT
        % ~\cite{tent_wang2020}
        & 92.80 / 92.80 & 76.88 / 79.08 & 56.72 / 65.96 & 50.45 / 58.45 \\
        % LAME~(CVPR22)
        LAME
        % ~\cite{niid_boudiaf2022parameter}
        & 40.50 / 40.50 & 43.66 / 44.88 & 44.15 / 46.64 & 43.81 / 40.33 \\
        % \change{COTTA}~(CVPR22)
        COTTA
        % ~\cite{cotta}
        & \change{52.20 / 52.20} & \change{44.48 / 40.93} & \change{45.46 / 38.77} & \change{45.67 / 33.72}\\
        % NOTE~(NIPS22)
        NOTE
        % ~\cite{note}
        & 73.80 / 73.80 & 57.71 / 58.86 & 54.44 / 57.10 & 53.74 / 52.48 \\
        % TTAC~(NIPS22)
        TTAC
        % ~\cite{su2022revisiting}
        & 34.10 / 34.10 & 40.48 / 38.28 & 47.84 / 41.47 & 49.78 / 38.00 \\
        % \change{PETAL}~(CVPR23)
        PETAL
        % ~\cite{brahma2023probabilistic}
        & \change{55.03 / 55.03} & \change{45.14 / 41.91} & \change{44.63 / 38.52} & \change{44.75 / 33.81}\\
        % RoTTA~(CVPR23)
        RoTTA
        % ~\cite{yuan2023robust}
        & 35.00 / 35.00 & 40.00 / 39.03 & 45.68 / 42.04 & 46.78 / 37.93 \\
        \midrule
        % \method & \bf 33.26\dtplus{+0.84} / \bf 33.26\dtplus{+0.84} & \bf 33.10\dtplus{+6.90} / \bf 34.31\dtplus{+3.97} & \bf 32.31\dtplus{+11.84} / \bf 34.98\dtplus{+3.54} & \bf 32.29\dtplus{11.52} / \bf 31.54\dtplus{+2.18} \\
        \method & \bf 33.26 / \bf 33.26 & \bf 33.10 / \bf 34.31 & \bf 32.31 / \bf 34.98 & \bf 32.29 / \bf 31.54 \\

        \midrule[1.2pt]
        \midrule[1.2pt]

        \multirow{2}{*}{Method} & \multicolumn{4}{c}{\textbf{Time-Varying Global Class Distribution (GLI-TTA-V)}}\\
        & $I.F.=1$ & $I.F.=10$ & $I.F.=100$ & $I.F.=200$  \\
        \midrule

        TEST  & 46.40 / 46.40 & 45.85 / 46.65 & 45.34 / 46.94 & 45.16 / 40.61\\
        BN
        % ~\cite{BN_Stat}
        & 52.90 / 52.90 & 45.10 / 42.47 & 45.15 / 38.80 & 45.37 / 33.45\\
        PL
        % ~\cite{PL}
        & 88.90 / 88.90 & 68.16 / 66.62 & 52.83 / 48.39 & 53.68 / 44.28\\
        % TENT~(ICLR21)
        TENT
        % ~\cite{tent_wang2020}
        & 92.80 / 92.80 & 77.11 / 76.51 & 65.42 / 63.48 & 62.45 / 53.57\\
        % LAME~(CVPR22)
        LAME
        % ~\cite{niid_boudiaf2022parameter}
        & 40.50 / 40.50 & 42.82 / 45.35 & 42.47 / 47.82 & 42.23 / 41.45\\
        % \change{COTTA}~(CVPR22)
        COTTA
        % ~\cite{cotta}
        & \change{52.20 / 52.20} & \change{43.74 / 41.03} & \change{43.83 / 37.93}  & \change{43.96 / 32.69}\\
        % NOTE~(NIPS22)
        NOTE
        % ~\cite{note}
        & 73.80 / 73.80 & 58.07 / 58.46 & 55.16 / 55.95 & 54.43 / 48.65\\
        % TTAC~(NIPS22)
        TTAC
        % ~\cite{su2022revisiting}
        & 34.10 / 34.10 & 38.56 / 38.68 & 42.07 / 41.05 & 42.87 / 35.80\\
        % \change{PETAL}~(CVPR23)
        PETAL
        % ~\cite{brahma2023probabilistic}
        & \change{55.03 / 55.03} & \change{44.36 / 41.54} & \change{44.11 / 38.33} & \change{44.43 / 32.84}\\
        % RoTTA~(CVPR23)
        RoTTA
        % ~\cite{yuan2023robust}
        & 35.00 / 35.00 & 39.56 / 39.77 & 42.20 / 39.93 & 43.57 / 35.82\\
        \midrule
        % \method & \bf 33.26\dtplus{+0.84} / \bf 33.26\dtplus{+0.84} & \bf 33.52\dtplus{+5.04} / \bf 34.49\dtplus{+4.19} & \bf 33.76\dtplus{+8.31} / \bf 34.63\dtplus{+3.30} & \bf 34.29\dtplus{+7.94} / \bf 30.22\dtplus{+2.47}\\
        \method & \bf 33.26 / \bf 33.26 & \bf 33.52 / \bf 34.49 & \bf 33.76 / \bf 34.63 & \bf 34.29 / \bf 30.22\\
        
        \bottomrule[1.2pt]
        \end{tabular}
    }
    % \vspace{-0.3cm}
    \label{tab:global_imbalance_cifar100}
\end{table}

\begin{table*}[t]
    \centering
    \caption{{Average classification error on ImageNet-C while continually adapting to different corruptions at the highest severity 5 with non-i.i.d. testing data stream.}
    % \vspace{-0.4cm}
    }
    % \vspace{-0.5cm}  
    \label{tab:ImageNet}
    \resizebox{0.95\linewidth}{!}{
    {
    \begin{tabular}{l|ccccccccccccccc|c}
        \toprule[1.2pt]
        Time & \multicolumn{15}{l|}{$t\xrightarrow{\hspace*{18.5cm}}$}& \\ \hline
        Method & \rotatebox[origin=c]{45}{motion} & \rotatebox[origin=c]{45}{snow} & \rotatebox[origin=c]{45}{fog} & \rotatebox[origin=c]{45}{shot} & \rotatebox[origin=c]{45}{defocus} & \rotatebox[origin=c]{45}{contrast} & \rotatebox[origin=c]{45}{zoom} & \rotatebox[origin=c]{45}{brightness} & \rotatebox[origin=c]{45}{frost} & \rotatebox[origin=c]{45}{elastic} & \rotatebox[origin=c]{45}{glass} & \rotatebox[origin=c]{45}{gaussian} & \rotatebox[origin=c]{45}{pixelate} & \rotatebox[origin=c]{45}{jpeg} & \rotatebox[origin=c]{45}{impulse}
        & Avg. \\ 
        
        \midrule
        TEST & 85.15 & 83.45 & 75.88 & 97.09 & 81.68 & 94.52 & 77.93 & 41.23 & 77.07 & 82.48 & 89.73 & 97.81 & 79.31 & 68.50 & 98.17 & 82.00 \\

        BN
        % ~\cite{BN_Stat}
        & 73.64 & 66.07 & 52.81 & 84.49 & 85.05 & 82.66 & 61.96 & 36.04 & 68.60 & 56.44 & 84.85 & 85.31 & 52.29 & 60.77 & 85.05 & 69.07 \\

        PL
        % ~\cite{PL}
        & 66.55 & 60.43 & 49.46 & \bf 76.57 & 79.23 & 81.04 & 65.35 & 51.48 & 75.62 & 69.74 & 89.04 & 92.36 & 86.84 & 92.09 & 97.83 & 75.58\\

        TENT
        % ~\cite{tent_wang2020}
        & \bf 64.37 & 59.73 & 51.20 & 77.47 & 81.70 & 88.72 & 82.38 & 76.91 & 93.64 & 95.43 & 98.80 & 98.98 & 98.39 & 98.90 & 99.40 & 84.40\\

        LAME
        % ~\cite{niid_boudiaf2022parameter}
        & 85.93 & 84.57 & 77.29 & 97.47 & 81.92 & 94.72 & 78.41 & 41.49 & 77.67 & 84.07 & 90.25 & 98.21 & 79.61 & 68.64 & 98.76 & 82.60\\
        EATA
        % ~\cite{niu2022efficient}
        & 73.15 & 65.41 & 52.51 & 84.27 & 85.09 & 82.85 & 61.52 & 35.15 & 68.26 & 56.30 & 84.43 & 84.95 & 51.63 & 60.85 & 85.05 & 68.76\\

        NOTE
        % ~\cite{note}
        & 82.97 & 78.29 & 73.43 & 93.92 & 96.35 & 89.73 & 93.18 & 84.57 & 92.82 & 94.54 & 98.50 & 98.88 & 98.17 & 97.55 & 98.78 & 91.44 \\
    
        ROTTA
        % ~\cite{yuan2023robust}
        & 74.86 & 70.02 & 55.26 & 85.55 & 85.37 & 78.61 & 61.00 & \bf 34.31 & 64.65 & 52.83 & 76.16 & 85.43 & \bf 48.70 & 52.41 & 78.37 & 66.90 \\
    
        \midrule
        % \method & 69.52 & \bf 59.55 & \bf 48.35 & 79.27 & \bf 78.47 & \bf 75.54 & \bf 56.62 & 35.19 & \bf 60.39 & \bf 49.26 & \bf 74.54 & \bf 74.10 & 50.08 & \bf 51.24 & \bf 72.59 & \bf 62.32\dtplus{+4.58} \\
        \method & 69.52 & \bf 59.55 & \bf 48.35 & 79.27 & \bf 78.47 & \bf 75.54 & \bf 56.62 & 35.19 & \bf 60.39 & \bf 49.26 & \bf 74.54 & \bf 74.10 & 50.08 & \bf 51.24 & \bf 72.59 & \bf 62.32 \\
    \bottomrule[1.2pt]
    \end{tabular}
    }
    }
    % \vspace{-0.3cm}
\end{table*}

\begin{figure*}
    \centering
    \includegraphics[width=0.95\linewidth]{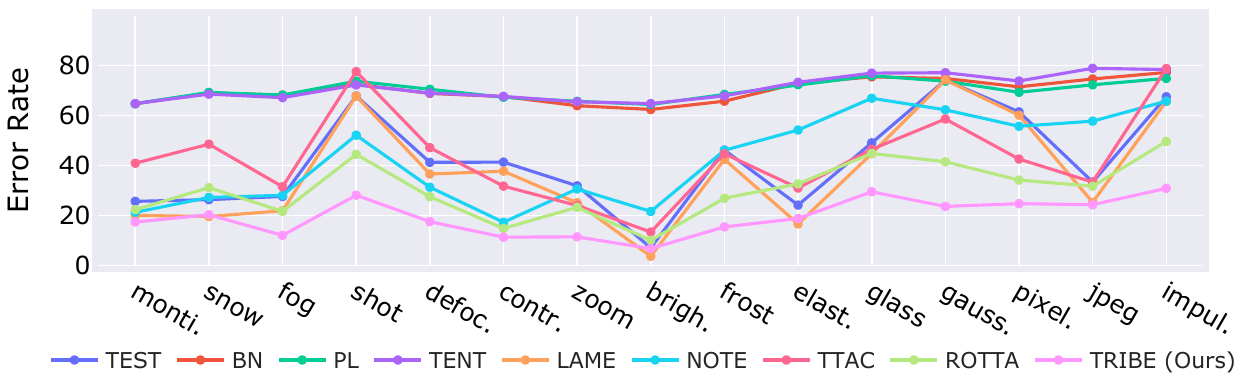}
    % \vspace{-0.2cm}
    \caption{{Performances on each individual domain (corruption) under GLI-TTA-F (I.F.=100) protocols on CIFAR10-C dataset.}
    % \vspace{-0.5cm}
    }
    % \vspace{-0.5cm}
    \label{fig:GLI-TTA-CIFAR10}
\end{figure*}

\noindent\textbf{Competing Methods}: We benchmark against the following TTA methods~\cite{BN_Stat, PL, tent_wang2020, niid_boudiaf2022parameter, note, su2022revisiting, yuan2023robust, niu2022efficient}. Direct testing (\textbf{TEST})  performs inference on test streaming data without adaptation. Prediction-time batch normalization (\textbf{BN})~\cite{BN_Stat} replaces the running statistics with the batch statistics on each testing minibatch for normalization. Pseudo Label (\textbf{PL})~\cite{PL} updates the parameters of all normalization layers by minimizing the cross-entropy loss with predicted pseudo labels. Test-time entropy minimization (\textbf{TENT})~\cite{tent_wang2020} updates the affine parameters of all batchnorm layers by minimizing the entropy of predictions. Laplacian adjusted maximum-likelihood estimation (\textbf{LAME})~\cite{niid_boudiaf2022parameter} adjusts the predictions of the model through maximizing the likelihood estimation without updating any parameters. 
\change{Continual test-time adaptation (\textbf{CoTTA})~\cite{cotta} performs mean-teacher architecture, and randomly selects and restores the parameters of the model to source model. \textbf{PETAL}~\cite{brahma2023probabilistic} leverages fisher information to instruct the parameter restoration.} 
Non-i.i.d. test-time adaptation (\textbf{NOTE})~\cite{note} optionally updates the batchnorm statistics when the distance between the instance statistics of the test sample and the source model's statistics is less than a threshold. Test-time anchored clustering (\textbf{TTAC})~\cite{su2022revisiting} minimizes the KL-Divergence between the source and target domain distributions. Robust test-time adaptation (\textbf{RoTTA})~\cite{yuan2023robust} replaces the batchnorm layers with Robust Batch Normalization for better estimation of target domain batchnorm statistics. Finally, we evaluate our \textbf{TRIBE} with tri-net self-training and Balanced Batchnorm layers.

% \vspace{-0.3cm}
\subsection{Real-World Test Time Adaptation Results}

Under the proposed real-world test-time adaptation protocol, the classification errors averaged over continuously adapting to all 15 types of corruptions under different degrees of global imbalance are calculated. We report the results in Tab.~\ref{tab:global_imbalance_cifar10} for CIFAR10-C and Tab.~\ref{tab:global_imbalance_cifar100} for CIFAR100-C. We make the following observations from the results. i) Direct testing without any adaptation is even stronger than many TTA methods. For example, only LAME, TTAC, RoTTA and our TRIBE could consistently outperform direct testing~(TEST) on both CIFAR10-C and CIFAR100-C datasets, suggesting the necessity to develop robust TTA approaches. ii) Global class imbalance poses a great challenge to existing robust TTA methods. For example, the previous state-of-the-art, RoTTA achieves $25.2\%$ and $35.0\%$ on CIFAR10-C and CIFAR100-C respectively, while the error rose to $30.04\%$ and $37.93\%$ under severely global imbalanced testing set ($I.F.=200$). The same observation applies to other competing methods. In comparison, TRIBE is able to maintain relatively better performance under more severe global imbalanced testing set. iii) We further notice that TRIBE consistently outperform all competing methods in absolute accuracy. Importantly, under balanced global distribution~($I.F.=1$), TRIBE outperforms the best performing model, TTAC, by $7\%$ on CIFAR10-C. The margin is maintained under more imbalanced testing set~($I.F.=200$). iv) TRIBE maintains a more consistent performance from $I.F.=10$ to $I.F.=200$ on both CIFAR10-C and CIFAR100-C, while other competing methods degenerate substantially. This is attributed to the introduction of Balanced BN layer better accounting for severe class imbalance and anchored loss avoiding over adaptation across the different domains.

\begin{table*}[t]
    \centering
    \caption{{Ablation study on CIFAR10/100-C under GLI-TTA-F ($I.F.=100$) protocol. We report classification error as evaluation metric. \change{MT* indicates Mean Teacher is adapted to TTA task by removing the labeled loss term.}
    % \vspace{-0.6cm}
    }
    }
    % \vspace{-0.5cm}
    \resizebox{0.93\linewidth}{!}{
        \begin{tabular}{c|cccc|cc|c}
        \toprule[1.2pt]
            Method & \change{EMA Model} & BatchNorm & Self-Training & Anchored Loss & CIFAR10-C & CIFAR100-C & Avg. \\
        \midrule
            TEST & -- & BN & -- & -- & 41.71 / 43.63 & 47.53 / 45.91 & 44.62 / 44.77 \\
            \change{ROTTA
            % ~\cite{yuan2023robust}
            }
            & \change{\checkmark} & \change{Robust BN} & \change{\checkmark} & \change{--} & \change{30.50 / 29.08} & \change{45.68 / 42.04} &  \change{38.09 / 35.56}\\
            -- & --& Robust BN & -- & -- & 43.48 / 32.29 & 40.45 / 36.94 & 41.97 / 34.62\\
            -- & -- & Balanced BN & -- & -- & 29.00 / 26.38 & 39.55 / 36.59 & 34.28 / 31.49 \\
            -- & -- & BN & \checkmark & -- & 37.67 / 38.94 & 37.12 / 44.77 & 37.40 / 41.86\\
            -- & -- & Balanced BN & \checkmark & -- & 36.58 / 65.88 & 37.21 / 44.83 & 36.90 / 55.36\\
            -- & -- & BN & \checkmark & \checkmark & 36.76 / 29.19 & 36.16 / 36.26 & 36.46 / 32.73\\
            \change{MT*} & \change{\checkmark} & \change{Balanced BN} & \change{\checkmark} & \change{--} & \change{23.76 / 25.18} & \change{36.01 / 35.72} & \change{29.89 / 30.45}\\
        \midrule
            \method & -- & Balanced BN & \checkmark & \checkmark & \bf 19.53 / \bf 24.66 & \bf 32.31 / \bf 34.98 & \bf 25.92 / \bf 29.82\\
        \bottomrule[1.2pt]
        \end{tabular}
    }    
    % \vspace{-0.3cm}
    \label{tab:component_ablation} 
\end{table*}

We further evaluate TTA performance on ImageNet-C dataset of which the testing set is naturally class imbalanced. Therefore, we only simulate local class imbalance for the testing data stream and allow $\alpha$ equal to the marginalized class distribution. We present both averaged and domain specific classification error in Tab.~\ref{tab:ImageNet}. We make similar observations with results on CIFAR10/100-C. Some competitive TTA methods perform exceptionally worse than direct testing while TRIBE again outperforms all competing methods both in terms of averaged error rate and winning on 11/15 corruption types.

\noindent\textbf{Results on Individual Corruption}: We adapt the model continually to constant shifting domains~(corruption types). We report the average classification error for each individual type of corruptions in Fig.~\ref{fig:GLI-TTA-CIFAR10}. We conclude from the plots that i) BN, PL and TENT normalize the features using the statistics calculated within current mini-batch, thus they all perform much worse than methods considering robust batchnorm e.g. NOTE, ROTTA and TRIBE. %and not updating any parameters e.g. LAME. 
ii) There is a strong correlation of performance across different methods suggesting certain corruptions, e.g. ``shot'', ``gaussian noise'' and ``impulse noise'', are inherently more difficult. Nevertheless, TRIBE always outperforms competing methods on these challenging corruptions. iii) Some competing methods achieve close to TRIBE accuracy on easier corruptions, but they often perform much worse on the upcoming corruptions. Overall, TRIBE exhibits much lower variance across all domains when continually adapted. This suggests the anchored loss potentially helps TRIBE to avoid over adapting to easier domains.

% \vspace{-0.2cm}
\subsection{Ablation \& Additional Study}
% \vspace{-0.1cm}

\noindent\textbf{Effect of Individual Components}: We investigate the effectiveness of proposed components in Tab.~\ref{tab:component_ablation}. Specifically, we first compare adaptation by updating batchnorm statistics. It is apparent that Balanced BN is substantially better than Robust BN~\cite{yuan2023robust} when separately applied. When a two branch self-training (teacher \& student net) is applied, we witness a clear improvement from the direct testing baseline. However the improvement is less significant by combining self-training with Balanced BN. This is probably caused by over adaptation to testing domains causing poor generalization to continually changing domains. This negative impact is finally remedied by introducing a tri-net architecture (Anchored Loss) which helps regularize self-training to avoid over adaptation.

\noindent\textbf{Comparing Batchnorm Layers}:
To evaluate the effectiveness of our proposed Balanced BN, we run forward pass for global and local class-imbalanced testing samples for multiple batch normalization modules proposed for real-world TTA, with results presented in Tab.~\ref{tab:BalancedBN_GLITTA}. We observe our proposed Balanced BN outperforms others with a large margin ($2.58 \sim 9.79\%$), especially under severely global class imbalance ($I.F.=200$). It further confirms that Balanced BN is more suitable for handling both global and local class-imbalanced testing data.

\begin{table}[!]
    \centering
    \resizebox{\linewidth}{!}{
        \begin{tabular}{l|cccc}
            \toprule[1.2pt]
            \multirow{2}{*}{Method} & \multicolumn{4}{c}{Imbalance Factor} \\
            & 1  & 10 & 100 & 200 \\
            \midrule
            TEST & 43.51 & 42.65 & 41.71 & 41.69 \\
            IABN~(NIPS22) & 27.29  & 31.74  & 38.19 & 40.03\\
            Robust BN (5e-2) & 46.33 & 39.49 & 43.48 & 45.39\\
            Robust BN (5e-3) & 29.50 & 36.38 & 44.36 & 46.71\\
            \bf Balanced BN (Ours) & \bf 24.71 & \bf 29.96 & \bf 29.00 & \bf 30.24\\
        \bottomrule[1.2pt]                
        \end{tabular}
        
    }    
    \caption{{The performance of different normalization layers which only updates the statistics. The classification error on CIFAR10-C are reported.}
    % \vspace{-0.3cm}
    } \label{tab:BalancedBN_GLITTA}
    % \vspace{-0.3cm}
\end{table}

\begin{figure}
    \centering
    % \vspace{-0.2cm}
    \includegraphics[width=\linewidth]{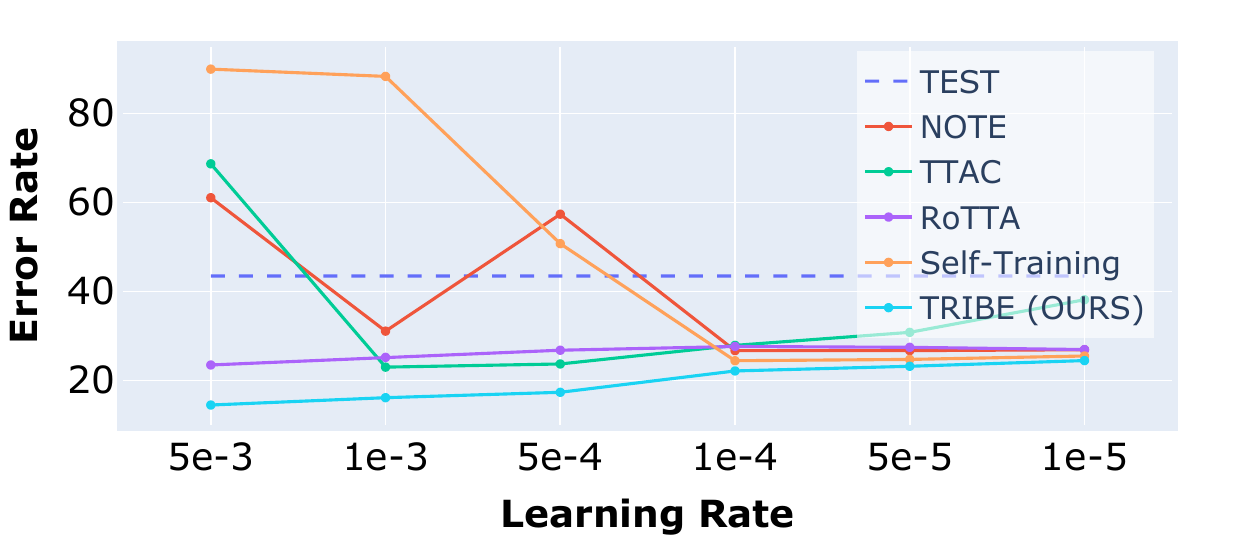}
    \caption{{We evaluate state-of-the-art TTA methods under different learning rates. The learning rates of NOTE fall in [5e-4, 1e-6] and TTAC fall in [5e-5, 1e-7] in order to align the best LR with other methods.}
    % \vspace{-0.6cm}
    }\label{fig:lr}
    % \vspace{-0.4cm}
\end{figure}

\noindent\textbf{Hyper-parameter Robustness}: Selecting appropriate hyper-parameter plays an important role in TTA~\cite{zhao2023pitfalls}. As TTA assumes no labeled data in testing set, selecting appropriate hyper-parameter becomes non-trivial. We argue that the tri-net design is naturally more robust to the choice of learning rate. As illustrated in Fig.~\ref{fig:lr}, TRIBE is very stable w.r.t. the choice of learning rate while other methods, e.g. TTAC and NOTE, prefer a much narrower range of learning rate. \change{More hyper-parameter analysis details can be found in the supplementary.}

\section{Conclusion}
% \vspace{-0.1cm}

In this work, we explore improving test-time adaptation algorithm's robustness to real-world challenges, including non-i.i.d. testing data stream, global class imbalance and continual domain shift. To adapt to imbalanced testing data, we propose a Balanced Batchnorm layer consisting of multiple category-wise statistics to achieve unbiased estimation of statistics. We further propose a tri-net architecture with student, teacher and anchor networks to regularize self-training based TTA. We demonstrate the effectiveness of the overall method, TRIBE, on simulated real-world test-time adaptation data streams. We achieve the state-of-the-art performance on all benchmarks created from four TTA datasets.

\noindent\textbf{Limitations}: TRIBE replaces regular Batchnorm layer with a customized Balanced Batchnorm layer, thus introducing additional storage overhead. Moreover, some recent Transformer based backbone network prefer Layernorm to Batchnorm~\cite{dosovitskiy2020vit}, thus potentially limiting the application of TRIBE. But recent studies revealed opportunities to integrate batchnorm to vision Transformer networks~\cite{yao2021leveraging}.

\section*{Acknowledgments}
This work is supported by National Natural Science Foundation of China (NSFC) (Grant Number: 62106078), and Agency for Science, Technology and Research (Grant Number: C210112059). This work was partially done during Yongyi Su's attachment with Institute for Infocomm Research (I2R), funded by China Scholarship Council~(CSC).

\bibliography{main}

\begin{thebibliography}{51}
\providecommand{\natexlab}[1]{#1}

\bibitem[{Arazo et~al.(2020)Arazo, Ortego, Albert, O’Connor, and
  McGuinness}]{arazo2020pseudo}
Arazo, E.; Ortego, D.; Albert, P.; O’Connor, N.~E.; and McGuinness, K. 2020.
\newblock Pseudo-labeling and confirmation bias in deep semi-supervised
  learning.
\newblock In \emph{International Joint Conference on Neural Networks}.

\bibitem[{Boudiaf et~al.(2022)Boudiaf, Mueller, Ben~Ayed, and
  Bertinetto}]{niid_boudiaf2022parameter}
Boudiaf, M.; Mueller, R.; Ben~Ayed, I.; and Bertinetto, L. 2022.
\newblock Parameter-free Online Test-time Adaptation.
\newblock In \emph{Proceedings of the IEEE/CVF Conference on Computer Vision
  and Pattern Recognition}.

\bibitem[{Brahma and Rai(2023)}]{brahma2023probabilistic}
Brahma, D.; and Rai, P. 2023.
\newblock A Probabilistic Framework for Lifelong Test-Time Adaptation.
\newblock In \emph{Proceedings of the IEEE/CVF Conference on Computer Vision
  and Pattern Recognition}, 3582--3591.

\bibitem[{Chen et~al.(2022)Chen, Wang, Darrell, and
  Ebrahimi}]{chen2022contrastive}
Chen, D.; Wang, D.; Darrell, T.; and Ebrahimi, S. 2022.
\newblock Contrastive Test-Time Adaptation.
\newblock In \emph{Proceedings of the IEEE/CVF Conference on Computer Vision
  and Pattern Recognition}.

\bibitem[{Croce et~al.(2021)Croce, Andriushchenko, Sehwag, Debenedetti,
  Flammarion, Chiang, Mittal, and Hein}]{croce2robustbench}
Croce, F.; Andriushchenko, M.; Sehwag, V.; Debenedetti, E.; Flammarion, N.;
  Chiang, M.; Mittal, P.; and Hein, M. 2021.
\newblock RobustBench: a standardized adversarial robustness benchmark.
\newblock In \emph{Thirty-fifth Conference on Neural Information Processing
  Systems Datasets and Benchmarks Track (Round 2)}.

\bibitem[{Cui et~al.(2019)Cui, Jia, Lin, Song, and Belongie}]{cui2019class}
Cui, Y.; Jia, M.; Lin, T.-Y.; Song, Y.; and Belongie, S. 2019.
\newblock Class-balanced loss based on effective number of samples.
\newblock In \emph{Proceedings of the IEEE/CVF Conference on Computer Vision
  and Pattern Recognition}.

\bibitem[{Dosovitskiy et~al.(2021)Dosovitskiy, Beyer, Kolesnikov, Weissenborn,
  Zhai, Unterthiner, Dehghani, Minderer, Heigold, Gelly, Uszkoreit, and
  Houlsby}]{dosovitskiy2020vit}
Dosovitskiy, A.; Beyer, L.; Kolesnikov, A.; Weissenborn, D.; Zhai, X.;
  Unterthiner, T.; Dehghani, M.; Minderer, M.; Heigold, G.; Gelly, S.;
  Uszkoreit, J.; and Houlsby, N. 2021.
\newblock An Image is Worth 16x16 Words: Transformers for Image Recognition at
  Scale.
\newblock In \emph{International Conference on Learning Representations}.

\bibitem[{Gandelsman et~al.(2022)Gandelsman, Sun, Chen, and
  Efros}]{gandelsman2022test}
Gandelsman, Y.; Sun, Y.; Chen, X.; and Efros, A.~A. 2022.
\newblock Test-time training with masked autoencoders.
\newblock In \emph{Advances in Neural Information Processing Systems}.

\bibitem[{Ganin and Lempitsky(2015)}]{ganin2015unsupervised}
Ganin, Y.; and Lempitsky, V. 2015.
\newblock Unsupervised domain adaptation by backpropagation.
\newblock In \emph{International Conference on Machine Learning}.

\bibitem[{Gong et~al.(2022)Gong, Jeong, Kim, Kim, Shin, and Lee}]{note}
Gong, T.; Jeong, J.; Kim, T.; Kim, Y.; Shin, J.; and Lee, S. 2022.
\newblock Robust Continual Test-time Adaptation: Instance-aware BN and
  Prediction-balanced Memory.
\newblock In \emph{Advances in Neural Information Processing Systems}.

\bibitem[{Goyal et~al.(2022)Goyal, Sun, Raghunathan, and
  Kolter}]{goyaltest2022}
Goyal, S.; Sun, M.; Raghunathan, A.; and Kolter, J.~Z. 2022.
\newblock Test Time Adaptation via Conjugate Pseudo-labels.
\newblock In \emph{Advances in Neural Information Processing Systems}.

\bibitem[{Gretton et~al.(2012)Gretton, Borgwardt, Rasch, Sch{\"o}lkopf, and
  Smola}]{gretton2012kernel}
Gretton, A.; Borgwardt, K.~M.; Rasch, M.~J.; Sch{\"o}lkopf, B.; and Smola, A.
  2012.
\newblock A kernel two-sample test.
\newblock \emph{Journal of Machine Learning Research}.

\bibitem[{He et~al.(2016)He, Zhang, Ren, and Sun}]{he2016deep}
He, K.; Zhang, X.; Ren, S.; and Sun, J. 2016.
\newblock Deep residual learning for image recognition.
\newblock In \emph{Proceedings of the IEEE/CVF Conference on Computer Vision
  and Pattern Recognition}, 770--778.

\bibitem[{Hendrycks and Dietterich(2019)}]{hendrycks2018benchmarking}
Hendrycks, D.; and Dietterich, T. 2019.
\newblock Benchmarking Neural Network Robustness to Common Corruptions and
  Perturbations.
\newblock In \emph{International Conference on Learning Representations}.

\bibitem[{Hoffman et~al.(2018)Hoffman, Tzeng, Park, Zhu, Isola, Saenko, Efros,
  and Darrell}]{hoffman2018cycada}
Hoffman, J.; Tzeng, E.; Park, T.; Zhu, J.-Y.; Isola, P.; Saenko, K.; Efros, A.;
  and Darrell, T. 2018.
\newblock Cycada: Cycle-consistent adversarial domain adaptation.
\newblock In \emph{International Conference on Machine Learning}.

\bibitem[{Ioffe and Szegedy(2015)}]{ioffe2015batch}
Ioffe, S.; and Szegedy, C. 2015.
\newblock Batch normalization: Accelerating deep network training by reducing
  internal covariate shift.
\newblock In \emph{International Conference on Machine Learning}.

\bibitem[{Iwasawa and Matsuo(2021)}]{iwasawa2021test}
Iwasawa, Y.; and Matsuo, Y. 2021.
\newblock Test-time classifier adjustment module for model-agnostic domain
  generalization.
\newblock \emph{Advances in Neural Information Processing Systems}, 34:
  2427--2440.

\bibitem[{Kingma and Ba(2014)}]{kingma2014adam}
Kingma, D.~P.; and Ba, J. 2014.
\newblock Adam: A method for stochastic optimization.
\newblock \emph{arXiv preprint arXiv:1412.6980}.

\bibitem[{Kumar, Ma, and Liang(2020)}]{kumar2020understanding}
Kumar, A.; Ma, T.; and Liang, P. 2020.
\newblock Understanding self-training for gradual domain adaptation.
\newblock In \emph{International Conference on Machine Learning}.

\bibitem[{Lecun et~al.(1998)Lecun, Bottou, Bengio, and Haffner}]{726791}
Lecun, Y.; Bottou, L.; Bengio, Y.; and Haffner, P. 1998.
\newblock Gradient-based learning applied to document recognition.
\newblock \emph{Proceedings of the IEEE}, 86(11): 2278--2324.

\bibitem[{Lee et~al.(2013)}]{PL}
Lee, D.-H.; et~al. 2013.
\newblock Pseudo-label: The simple and efficient semi-supervised learning
  method for deep neural networks.
\newblock In \emph{Workshop on challenges in representation learning, ICML},
  volume~3, 896.

\bibitem[{Lim et~al.(2023)Lim, Kim, Choo, and Choi}]{lim2023ttn}
Lim, H.; Kim, B.; Choo, J.; and Choi, S. 2023.
\newblock TTN: A domain-shift aware batch normalization in test-time
  adaptation.
\newblock \emph{arXiv preprint arXiv:2302.05155}.

\bibitem[{Liu, Wang, and Long(2021)}]{liu2021cycle}
Liu, H.; Wang, J.; and Long, M. 2021.
\newblock Cycle self-training for domain adaptation.
\newblock In \emph{Advances in Neural Information Processing Systems}.

\bibitem[{Liu et~al.(2021)Liu, Kothari, van Delft, Bellot-Gurlet, Mordan, and
  Alahi}]{liu2021ttt++}
Liu, Y.; Kothari, P.; van Delft, B.; Bellot-Gurlet, B.; Mordan, T.; and Alahi,
  A. 2021.
\newblock TTT++: When Does Self-Supervised Test-Time Training Fail or Thrive?
\newblock In \emph{Advances in Neural Information Processing Systems}.

\bibitem[{Long et~al.(2015)Long, Cao, Wang, and Jordan}]{long2015learning}
Long, M.; Cao, Y.; Wang, J.; and Jordan, M. 2015.
\newblock Learning transferable features with deep adaptation networks.
\newblock In \emph{International Conference on Machine Learning}.

\bibitem[{Loshchilov and Hutter()}]{loshchilovsgdr}
Loshchilov, I.; and Hutter, F. ????
\newblock SGDR: Stochastic Gradient Descent with Warm Restarts.
\newblock In \emph{International Conference on Learning Representations}.

\bibitem[{Mu and Gilmer(2019)}]{mu2019mnist}
Mu, N.; and Gilmer, J. 2019.
\newblock Mnist-c: A robustness benchmark for computer vision.
\newblock \emph{arXiv preprint arXiv:1906.02337}.

\bibitem[{Nado et~al.(2020)Nado, Padhy, Sculley, D'Amour, Lakshminarayanan, and
  Snoek}]{BN_Stat}
Nado, Z.; Padhy, S.; Sculley, D.; D'Amour, A.; Lakshminarayanan, B.; and Snoek,
  J. 2020.
\newblock Evaluating Prediction-Time Batch Normalization for Robustness under
  Covariate Shift.
\newblock \emph{CoRR}, abs/2006.10963.

\bibitem[{Niu et~al.(2022)Niu, Wu, Zhang, Chen, Zheng, Zhao, and
  Tan}]{niu2022efficient}
Niu, S.; Wu, J.; Zhang, Y.; Chen, Y.; Zheng, S.; Zhao, P.; and Tan, M. 2022.
\newblock Efficient test-time model adaptation without forgetting.
\newblock In \emph{International Conference on Machine Learning}.

\bibitem[{Niu et~al.(2023)Niu, Wu, Zhang, Wen, Chen, Zhao, and
  Tan}]{niu2023towards}
Niu, S.; Wu, J.; Zhang, Y.; Wen, Z.; Chen, Y.; Zhao, P.; and Tan, M. 2023.
\newblock Towards stable test-time adaptation in dynamic wild world.
\newblock \emph{International Conference on Learning Representations}.

\bibitem[{Sakaridis, Dai, and Van~Gool(2018)}]{sakaridis2018semantic}
Sakaridis, C.; Dai, D.; and Van~Gool, L. 2018.
\newblock Semantic foggy scene understanding with synthetic data.
\newblock \emph{International Journal of Computer Vision}.

\bibitem[{Sohn et~al.(2020)Sohn, Berthelot, Carlini, Zhang, Zhang, Raffel,
  Cubuk, Kurakin, and Li}]{sohn2020fixmatch}
Sohn, K.; Berthelot, D.; Carlini, N.; Zhang, Z.; Zhang, H.; Raffel, C.~A.;
  Cubuk, E.~D.; Kurakin, A.; and Li, C.-L. 2020.
\newblock Fixmatch: Simplifying semi-supervised learning with consistency and
  confidence.
\newblock \emph{Advances in Neural Information Processing Systems}.

\bibitem[{Su, Xu, and Jia(2022)}]{su2022revisiting}
Su, Y.; Xu, X.; and Jia, K. 2022.
\newblock Revisiting Realistic Test-Time Training: Sequential Inference and
  Adaptation by Anchored Clustering.
\newblock In \emph{Advances in Neural Information Processing Systems}.

\bibitem[{Su et~al.(2023)Su, Xu, Li, and Jia}]{su2023revisiting}
Su, Y.; Xu, X.; Li, T.; and Jia, K. 2023.
\newblock Revisiting Realistic Test-Time Training: Sequential Inference and
  Adaptation by Anchored Clustering Regularized Self-Training.
\newblock \emph{arXiv preprint arXiv:2303.10856}.

\bibitem[{Sun and Saenko(2016)}]{sun2016deep}
Sun, B.; and Saenko, K. 2016.
\newblock Deep coral: Correlation alignment for deep domain adaptation.
\newblock In \emph{European Conference on Computer Vision}.

\bibitem[{Sun et~al.(2022)Sun, Zhang, Kailkhura, Yu, Xiao, and
  Mao}]{ModelNet40-C}
Sun, J.; Zhang, Q.; Kailkhura, B.; Yu, Z.; Xiao, C.; and Mao, Z.~M. 2022.
\newblock Benchmarking Robustness of 3D Point Cloud Recognition Against Common
  Corruptions.
\newblock \emph{arXiv preprint arXiv:2201.12296}.

\bibitem[{Sun et~al.(2020)Sun, Wang, Liu, Miller, Efros, and
  Hardt}]{sun2020test}
Sun, Y.; Wang, X.; Liu, Z.; Miller, J.; Efros, A.; and Hardt, M. 2020.
\newblock Test-time training with self-supervision for generalization under
  distribution shifts.
\newblock In \emph{International Conference on Machine Learning}.

\bibitem[{Tang, Chen, and Jia(2020)}]{tang2020unsupervised}
Tang, H.; Chen, K.; and Jia, K. 2020.
\newblock Unsupervised domain adaptation via structurally regularized deep
  clustering.
\newblock In \emph{Proceedings of the IEEE/CVF Conference on Computer Vision
  and Pattern Recognition}.

\bibitem[{Tzeng et~al.(2014)Tzeng, Hoffman, Zhang, Saenko, and
  Darrell}]{tzeng2014deep}
Tzeng, E.; Hoffman, J.; Zhang, N.; Saenko, K.; and Darrell, T. 2014.
\newblock Deep domain confusion: Maximizing for domain invariance.
\newblock \emph{arXiv preprint arXiv:1412.3474}.

\bibitem[{Wang et~al.(2021)Wang, Shelhamer, Liu, Olshausen, and
  Darrell}]{tent_wang2020}
Wang, D.; Shelhamer, E.; Liu, S.; Olshausen, B.~A.; and Darrell, T. 2021.
\newblock Tent: Fully Test-Time Adaptation by Entropy Minimization.
\newblock In \emph{International Conference on Learning Representations}.

\bibitem[{Wang et~al.(2022{\natexlab{a}})Wang, Lan, Liu, Ouyang, Qin, Lu, Chen,
  Zeng, and Yu}]{wang2022generalizing}
Wang, J.; Lan, C.; Liu, C.; Ouyang, Y.; Qin, T.; Lu, W.; Chen, Y.; Zeng, W.;
  and Yu, P. 2022{\natexlab{a}}.
\newblock Generalizing to unseen domains: A survey on domain generalization.
\newblock \emph{IEEE Transactions on Knowledge and Data Engineering}.

\bibitem[{Wang and Deng(2018)}]{wang2018deep}
Wang, M.; and Deng, W. 2018.
\newblock Deep visual domain adaptation: A survey.
\newblock \emph{Neurocomputing}.

\bibitem[{Wang et~al.(2022{\natexlab{b}})Wang, Fink, Gool, and Dai}]{cotta}
Wang, Q.; Fink, O.; Gool, L.~V.; and Dai, D. 2022{\natexlab{b}}.
\newblock Continual Test-Time Domain Adaptation.
\newblock In \emph{Proceedings of the IEEE/CVF Conference on Computer Vision
  and Pattern Recognition}.

\bibitem[{Wu et~al.(2021)Wu, Guo, Su, and Weinberger}]{wu2021online}
Wu, R.; Guo, C.; Su, Y.; and Weinberger, K.~Q. 2021.
\newblock Online adaptation to label distribution shift.
\newblock In \emph{Advances in Neural Information Processing Systems}.

\bibitem[{Xie et~al.(2017)Xie, Girshick, Doll{\'a}r, Tu, and
  He}]{xie2017aggregated}
Xie, S.; Girshick, R.; Doll{\'a}r, P.; Tu, Z.; and He, K. 2017.
\newblock Aggregated residual transformations for deep neural networks.
\newblock In \emph{Proceedings of the IEEE/CVF Conference on Computer Vision
  and Pattern Recognition}, 1492--1500.

\bibitem[{Yao et~al.(2021)Yao, Cao, Lin, Liu, Zhang, and
  Hu}]{yao2021leveraging}
Yao, Z.; Cao, Y.; Lin, Y.; Liu, Z.; Zhang, Z.; and Hu, H. 2021.
\newblock Leveraging batch normalization for vision transformers.
\newblock In \emph{Proceedings of the IEEE International Conference on Computer
  Vision}.

\bibitem[{Yuan, Xie, and Li(2023)}]{yuan2023robust}
Yuan, L.; Xie, B.; and Li, S. 2023.
\newblock Robust Test-Time Adaptation in Dynamic Scenarios.
\newblock In \emph{Proceedings of the IEEE/CVF Conference on Computer Vision
  and Pattern Recognition}.

\bibitem[{Zagoruyko and Komodakis(2016)}]{zagoruyko2016wide}
Zagoruyko, S.; and Komodakis, N. 2016.
\newblock Wide Residual Networks.
\newblock In \emph{British Machine Vision Conference 2016}. British Machine
  Vision Association.

\bibitem[{Zellinger et~al.(2016)Zellinger, Grubinger, Lughofer,
  Natschl{\"a}ger, and Saminger-Platz}]{zellinger2017central}
Zellinger, W.; Grubinger, T.; Lughofer, E.; Natschl{\"a}ger, T.; and
  Saminger-Platz, S. 2016.
\newblock Central moment discrepancy (cmd) for domain-invariant representation
  learning.
\newblock In \emph{International Conference on Learning Representations}.

\bibitem[{Zhang et~al.(2022)Zhang, Hooi, Hong, and Feng}]{zhang2022self}
Zhang, Y.; Hooi, B.; Hong, L.; and Feng, J. 2022.
\newblock Self-supervised aggregation of diverse experts for test-agnostic
  long-tailed recognition.
\newblock In \emph{Advances in Neural Information Processing Systems}.

\bibitem[{Zhao et~al.(2023)Zhao, Liu, Alahi, and Lin}]{zhao2023pitfalls}
Zhao, H.; Liu, Y.; Alahi, A.; and Lin, T. 2023.
\newblock On Pitfalls of Test-Time Adaptation.
\newblock In \emph{ICLR 2023 Workshop on Pitfalls of limited data and
  computation for Trustworthy ML}.

\end{thebibliography}

\newpage

\appendix

% Alternative with a box around the title
% \noindent\fbox{%
% \begin{minipage}{\dimexpr\textwidth-2\fboxsep-2\fboxrule\relax}
% \begin{center}
% \Large\bfseries Appendix of "Towards Real-World Test-Time Adaptation: Tri-Net Self-Training with Balanced Normalization"
% \end{center}
% \end{minipage}}
% \vspace{1em}

\twocolumn[{
\vbox{%
    \hsize\textwidth%
    \linewidth\hsize%
    \centering%
    {\LARGE\bf Appendix of "Towards Real-World Test-Time Adaptation: Tri-Net Self-Training with Balanced Normalization" \par}%
    \vskip 0.625in minus 0.125in%
}%
}]

\appendix

In this supplementary material, we first review the real-world test-time adaptation evaluation protocols and highlight the challenges arising from the proposed TTA protocol. We further present evaluations on MNIST-C dataset and demonstrate that TRIBE achieves the state-of-the-art performance. Next, we provide more in-depth evaluations into individual components, more fine-grained evaluations of different degrees of class-imbalance and detailed evaluations of sensitivity to hyper-parameters. Related works about test-time strategies are discussed. Eventually, we present a derivation for the proposed Balanced BN layer, details for data augmentation adopted in self-training and more detailed results on each type of corruption across all datasets.

% \noindent\textbf{}
% \section{Comparison between several existing TTA protocols and our proposed GLI-TTA protocols}

\section{Details of real-world TTA protocols}

We aim to simulate a real-world test-time adaptation protocol by considering three key factors, namely non-stationary domain shift, imbalanced local class distribution and imbalanced global class distribution. First of all, we compare the challenging real-world TTA protocol against existing evaluation protocols in Tab.~\ref{tab:SettingComparison}. The ``Fully Test-Time Adaptation''~\cite{tent_wang2020,su2022revisiting} makes the strongest assumption on the testing data stream by assuming a static testing domain shift and balanced class distribution both locally and globally. ``Continual Test-Time Adaptation''~\cite{cotta} further advances towards a more realistic TTA protocol by allowing testing domain to shift over time. ``Non-i.i.d. Test-Time Adaptation''~\cite{note} further proposed a protocol simulating the testing data stream drawn in a non-i.i.d. manner from different classes, resulting imbalanced class distribution within a local time window. ``Pratical Test-Time Adaptation''~\cite{yuan2023robust} combines the above two real challenges together which poses more challenges to existing TTA methods. Finally, we would like to consider additionally that the global class distribution on the testing data stream could be imbalanced and introduce two variants of real-world TTA protocols for Global and Local class Imbalanced Test Time Adaptation~(GLI-TTA), namely 
 ``GLI-TTA-F'' and ``GLI-TTA-V'', to mimic the most difficult testing data stream. In particular, ``GLI-TTA-F'' assumes the global class distribution is fixed over the whole testing data stream while ``GLI-TTA-V'' allows the global class distribution to vary over time. The more specific comparisons of different TTA protocols are presented in Tab.~\ref{tab:SettingComparison}.

\begin{figure}[t]
    \centering
    
    \includegraphics[width=0.8\linewidth]{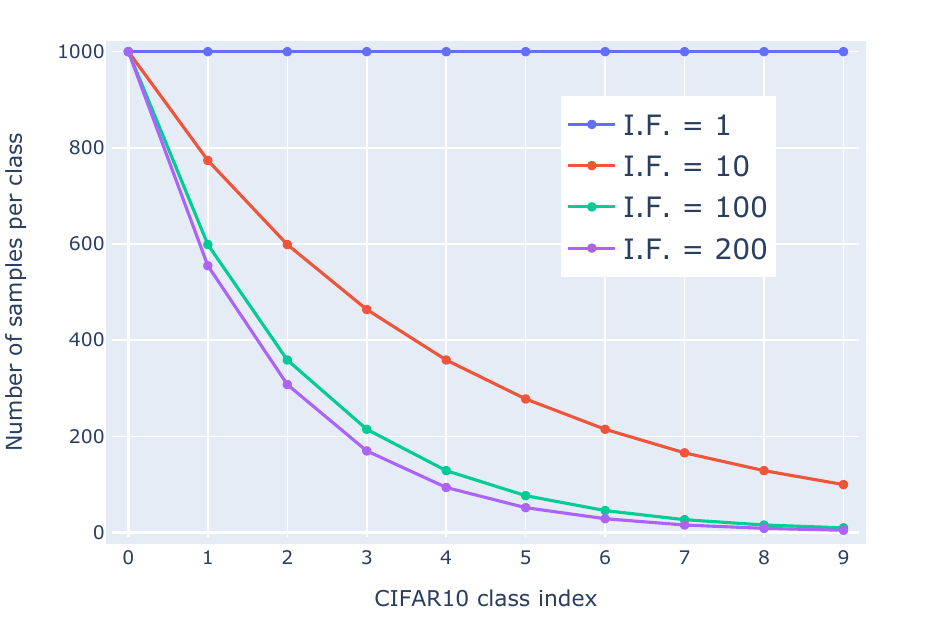}
    \vspace{-0.3cm}
    \caption{The number of testing samples per class in each domain (type of corruption) on CIFAR10-C under different imbalance degrees of GLI-TTA protocols.}\label{fig:ClassDistCurve_C10}
\end{figure}

\begin{figure}[h]
    \centering
    \vspace{-0.5cm}
    
    \includegraphics[width=0.8\linewidth]
    {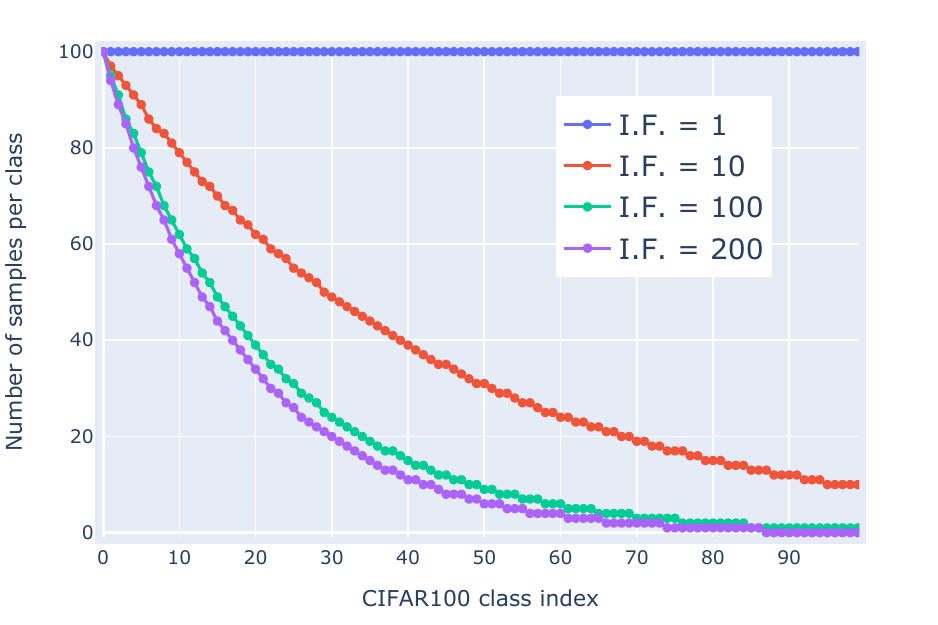}
    \vspace{-0.3cm}
    \caption{The number of testing samples per class in each domain (type of corruption) on CIFAR100-C under different imbalance degrees of GLI-TTA protocols.}\label{fig:ClassDistCurve_C100}
\end{figure}

\noindent\textbf{Global Class-Imbalance}

In this section, we show the class distributions under different GLI-TTA protocols on CIFAR10-C and CIFAR100-C in Fig.~\ref{fig:ClassDistCurve_C10} and Fig.~\ref{fig:ClassDistCurve_C100} respectively. We follow the strategy proposed in \cite{cui2019class} to generate imbalanced testing data stream. Specifically, we define the proportion parameter as $\alpha_k=\sigma\cdot (\frac{1}{I.F.})^{k/K_c}$, where $\sigma$ controls the degree of local class-imbalance, $I.F.$ denotes the imbalanced factor and $k$ refers to the index of semantic category.

\begin{table*}[!ht]
    \centering
    \caption{Comparing existing real-world TTA protocols with our proposed GLI-TTA protocols}
    
    \resizebox{0.9\textwidth}{!}{
        \begin{tabular}{l|ccc}
        \toprule[1.2pt]
           Setting  & Stationary Domain & Balanced Local Class Distribution & Global Class Distribution \\
        \midrule
           Fully Test-Time Adaptation~\cite{tent_wang2020}  & \Checkmark & \Checkmark & Balanced \\
           Continual Test-Time Adaptation~\cite{cotta} & \XSolidBrush & \Checkmark & Balanced \\
           Non-i.i.d. Test-Time Adaptation~\cite{note} & \Checkmark & \XSolidBrush & Balanced \\
           Practical Test-Time Adaptation~\cite{yuan2023robust} & \XSolidBrush & \XSolidBrush & Balanced \\
        \midrule
            \bf GLI-TTA-F (Ours) & \XSolidBrush & \XSolidBrush & Imbalanced \& Fixed \\
            \bf GLI-TTA-V (Ours) & \XSolidBrush & \XSolidBrush & Imbalanced \& Varying \\
        \bottomrule[1.2pt]
        \end{tabular}
    }
    \label{tab:SettingComparison}
\end{table*}

\begin{table*}[h]
    \centering
    \caption{\footnotesize{Average classification error (lower the better) on MNIST-C while continually adapting to different corruptions with globally and locally class-imbalanced test stream. $I.F.$ is the Imbalance Factor of Global Class-Imbalance. Instance-wise average error rate $a\%$ and  category-wise average error rate $b\%$ are separated by (a / b).}}
    \resizebox{0.85\textwidth}{!}{
        \begin{tabular}{l|c|c|c|c}
        \toprule[1.2pt]
        \multirow{2}{*}{Method} & \multicolumn{4}{c}{\textbf{Fixed Global
        Class Distribution (GLI-TTA-F)}}\\
        & $I.F.=1$ & $I.F.=10$ & $I.F.=100$ & $I.F.=200$ \\
        \midrule
        TEST  & 14.52 / 14.53 & 15.10 / 14.57 & 15.63 / 14.64 & 15.73 / 14.25\\
        BN~\cite{BN_Stat}    &  63.76 / 63.75 & 61.27 / 54.48 & 61.15 / 39.31 & 61.42 / 35.27\\
        PL~\cite{PL}    & 67.58 / 67.55 & 62.65 / 55.89 & 61.94 / 40.35 & 60.73 / 35.83\\
        TENT~\cite{tent_wang2020}  & 70.60 / 70.55 & 64.51 / 57.91 & 61.12 / 40.91 & 60.53 / 35.80 \\
        LAME~\cite{niid_boudiaf2022parameter}  & 12.69 / 12.70 & 13.44 / 12.81 & 14.07 / 13.88 & 14.25 / 14.71 \\
        NOTE~\cite{note}  & 10.08 / 10.04 & 11.73 / 9.92 & 15.02 / 10.50 & 14.67 / 10.06 \\
        TTAC~\cite{su2022revisiting}  & 13.02 / 13.03 & 15.10 / 14.48 & 15.43 / 14.60 & 15.48 / 14.20 \\
        RoTTA~\cite{yuan2023robust} & 7.62 / 7.60 & 8.94 / 8.41 & 11.28 / 10.36 & 11.31 / 9.53 \\
        \midrule
        \method & \bf 7.60 / \bf 7.58 & \bf 8.11 / \bf 8.11 & \bf 8.30 / \bf 8.02 & \bf 8.22 / \bf 7.14\\

        \midrule[1.2pt]
        \midrule[1.2pt]

        \multirow{2}{*}{Method} & \multicolumn{4}{c}{\textbf{Time-Varying Global Class Distribution (GLI-TTA-V)}}\\
        & $I.F.=1$ & $I.F.=10$ & $I.F.=100$ & $I.F.=200$  \\
        \midrule
        TEST  & 14.52 / 14.53 & 12.42 / 14.68 & 10.60 / 14.13 & 10.31 / 14.07\\
        BN~\cite{BN_Stat}    & 63.76 / 63.75 & 58.65 / 53.88 & 56.79 / 37.25 & 57.08 / 34.25 \\
        PL~\cite{PL}    & 67.58 / 67.55 & 61.05 / 55.95 & 58.35 / 38.45 & 58.06 / 34.62 \\
        TENT~\cite{tent_wang2020}  & 70.60 / 70.55 & 62.00 / 57.33 & 59.10 / 39.18 & 59.28 / 36.00\\
        LAME~\cite{niid_boudiaf2022parameter}  & 12.69 / 12.70 & 10.47 / 12.94 & 8.69 / 13.46 & \textbf{8.25} / 13.87\\
        NOTE~\cite{note}  & 10.08 / 10.04 & 9.30 / 9.15 & 9.18 / 9.02 & 9.01 / 8.88\\
        TTAC~\cite{su2022revisiting}  & 13.02 / 13.03 & 12.15 / 14.32 & 10.46 / 14.09 & 10.18 / 14.09\\
        RoTTA~\cite{yuan2023robust} & 7.62 / 7.60 & 8.45 / 8.62 & 9.33 / 9.22 & 9.74 / 9.48\\
        \midrule
        \method & \bf 7.60 / \bf 7.58 & \bf 8.28 / \bf 8.11 & \bf 8.28 / \bf 7.39 & 8.77 / \bf 8.01\\
        
        \bottomrule[1.2pt]
        \end{tabular}
    }
    \label{tab:global_imbalance_mnist}
\end{table*}

\section{Additional Evaluation}

\subsection{Evaluation on MNIST-C}

MNIST-C~\cite{mu2019mnist} is constructed by injecting simulated corruptions to the testing set of MNIST~\cite{726791}. MNIST-C includes $15$ types of corruptions to the digit images, consisting of Shot Noise, Impulse Noise, Glass Blur, Motion Blur, Shear, Scale, Rotate, Brightness, Translate, Stripe, Fog, Spatter, Dotted Line, Zigzag, and Canny Edges. Each corruption set has $10,000$ testing samples divided into $10$ categories. We train a ResNet18~\cite{he2016deep} on clean MNIST training set for $100$ epochs, using Adam~\cite{kingma2014adam} with initialized learning rate $1e-3$ and cosine annealing learning rate scheduling~\cite{loshchilovsgdr}. The pre-trained model is used for test-time adaptation on MNIST-C testing set under GLI-TTA protocols. To simulate continually shifting domains, we adopt a domain (corruption) shifting order as "$Shot Noise \to Impulse Noise \to Glass Blur \to Motion Blur \to Shear \to Scale \to Rotate \to Brightness \to Translate \to Stripe \to Fog \to Spatter \to Dotted Line \to Zigzag \to Canny Edges$". Quantitative results are shown in Tab.~\ref{tab:global_imbalance_mnist}. We make the similar observations with the results on CIFAR10/100-C and ImageNet-C evaluated in the manuscript as below. i) Some test-time adaptations (e.g., BN, PL and TENT) obtain worse results than without adaptation, which demonstrates it's necessary to propose a more robust test-time adaptation method which can better handle different kinds of test streams. ii) Our proposed \textbf{TRIBE} consistently outperforms other competing methods in both challenging GLI-TTA protocols.

\begin{table*}[t]
    \centering
    \caption{\change{Analysis of individual components. Consis. Const. indicates Consistency Constraint. The \textbf{bold} text indicates our contributed module. Mean Teacher* and FixMatch* are adapted to TTA task by removing the labeled loss term.}}
    \label{tab:ComponentAnlysis}
    \resizebox{\textwidth}{!}{
        \begin{tabular}{c|c|cccccc|c}
        \toprule[1.2pt]
            \multirow{2.5}{*}{Line} & \multirow{2.5}{*}{Method} & \multicolumn{6}{c|}{Component} & \multirow{2.5}{*}{Avg. Error Rate} \\
            \cmidrule{3-8}
             & & EMA Model & Category-balanced Memory~\cite{yuan2023robust} & Robust BN~\cite{yuan2023robust} & \textbf{Balanced BN} & Consis. Const. / ST Loss & \textbf{Anchored Loss} & \\
        \midrule
            1 & TEST & -- & -- & -- & -- & -- & -- & 43.5 \\
        \midrule
            \rowcolor{pink} 2 & RoTTA~\cite{yuan2023robust} & \checkmark & \checkmark & \checkmark & -- & \checkmark & -- & 25.2 \\
            \rowcolor{olive} 3 & -- & -- & \checkmark & \checkmark & -- & \checkmark & -- & 58.5\\
            \rowcolor{olive} 4 & -- & -- & \checkmark & \checkmark & -- & \checkmark & \checkmark & 25.5\\
            5 & -- & -- & \checkmark & -- & \checkmark & \checkmark & \checkmark & 25.0\\
            6 & -- & -- & -- & \checkmark & -- & \checkmark & -- & 46.0 \\
            7 & -- & -- & -- & \checkmark & -- & \checkmark & \checkmark & 40.3\\
            8 & -- & \checkmark & -- & \checkmark & -- & \checkmark & -- & 44.1 \\
            \rowcolor{pink} 9 & Mean Teacher* & \checkmark & -- & -- & \checkmark & \checkmark & -- & \underline{21.6} \\
            10 & -- & \checkmark & \checkmark & -- & \checkmark & \checkmark & -- & 23.3 \\
            11 & FixMatch* & -- & -- & -- & \checkmark & \checkmark & -- & 88.4\\
            12 & -- & -- & -- & -- & -- & \checkmark & \checkmark & 34.9 \\
        \midrule
            13 & Robust BN~\cite{yuan2023robust} & -- & -- & \checkmark & -- & -- & -- & 46.3 \\
            14 & \bf Balanced BN & -- & -- & -- & \checkmark & -- & -- & 24.7 \\
        \midrule
            15 & \method & -- & -- & -- & \checkmark & \checkmark & \checkmark & \bf 16.1 \\
        \bottomrule[1.2pt]
        \end{tabular}
    }
\end{table*}

\subsection{The effectiveness of individual components under non-i.i.d. TTA protocol}

In this section, we provide an in-depth comparative study of the individual components used in RoTTA~\cite{yuan2023robust} and our proposed \textbf{TRIBE}. We follow the PTTA protocol defined in \cite{yuan2023robust} by evaluating on CIFAR10-C with local class-imbalance and continual domain shift. Specifically, we analyze the components including i) exponentially moving average model (EMA Model), ii) category-balanced memory~\cite{yuan2023robust}, iii) Robust BN~\cite{yuan2023robust}, iv) Balanced BN, v) consistency constraint or self-training loss (Consis. Const. / ST Loss) and vi) anchored loss. From the results shown in Tab.~\ref{tab:ComponentAnlysis}, we observe that, i) comparing lines $2$ and $9$, our proposed Balanced BN outperforms the combination of category-balanced memory and Robust BN ($25.2\% \to 21.6\%$), which demonstrates the effectiveness of our proposed Balanced BN under local class-imbalance. Although RoTTA tries to smooth the impact from non-i.i.d. data through memory bank and robust batchnorm, memory bank highly depends on the pseudo labels and the label-inaccurate samples would be accumulated in the memory. In our proposed \textbf{TRIBE}, we abandon the memory bank and adapt the non-i.i.d. test streaming data by our proposed Balanced BN, which demonstrates strong capability in adapting to locally imbalanced testing data stream. ii) Compared with line $3$, the approach with our proposed Anchored Loss (line $4$) achieves better performance. The similar observation, e.g., lines $6$ and $7$, also demonstrates the effectiveness of Anchored Loss. iii) In addition to the positive results above, we observe only combining Balanced BN and ST Loss (line $11$) causes a bad result. We hypothesize that it is due to the confirmation bias accumulation and over-adaptation caused by only performing ST Loss without Anchored Loss and this gradually severe confirmation bias in turn affects the Balanced BN. Fortunately, this negative impact is finally remedied by introducing Anchored Loss (\textbf{TRIBE} line 15).

\subsection{Evaluation on gradually changing corruptions.}

\change{To simulate such a non-stationary corruption severity we carried out an experiment by repeating TTA on the testing data stream 5 times with an increasing corruption severity from 1 to 5. Both local and global class-imbalance, GLI-TTA-F (IF=100), are simulated in this scenario. As shown in Tab.~\ref{tab:graudallychanging}, TRIBE still outperforms all competing methods with a large margin under non-stationary corruption severity.}

\begin{table}[h]
    \centering
    \resizebox{\linewidth}{!}{
        \begin{tabular}{c|cccccc}
        \toprule[1.2pt]
            Avg Error & TEST  &  BN   & TENT  & CoTTA & RoTTA &   TRIBE \\
        \midrule
            CIFAR10C & 22.89 & 67.31 & 74.75 & 75.78 & 20.18 & \bf 10.49 \\
        \bottomrule[1.2pt]
        \end{tabular}
    }
    \caption{TTA performance on CIFAR10-C under gradually changing corruption severity. Both local and global class-imbalance, i.e. GLI-TTA-F (IF=100), are simulated in the experiment.}
    \label{tab:graudallychanging}
\end{table}

\subsection{Computation Cost Measured in Wall-Clock Time}

We evaluate the computation cost of our proposed TRIBE in this section. It's worthy note that, we only update the weight and bias parameters in batchnorm layers so that most of parameters across three models are shared, with only a small fraction of the remaining independent parameters. On the other hand, we have implemented Balanced BN as efficiently as possible using cpp in order to make the algorithm more efficient and more effective. Here, we provide the computation cost analysis of our TRIBE and several SOTA TTA methods in Tab.~\ref{tab:wall_clock}. We make the following observations. Our model undeniably got the best result and by a huge margin with others, and our model didn't take too much extra time compared to other SOTAs (only 1ms more than ROTTA), which we feel is acceptable for a TTA method.

\begin{table}[h]
    \centering
    \resizebox{\linewidth}{!}{
        \begin{tabular}{c|c|c}
        \toprule[1.2pt]
           Method  &  Error Rate & Inference and Adaptation Time per Sample (second)\\
        \midrule
           TEST  & 41.71 & 0.0005 \\
           BN    & 70.00 & 0.0005 \\
           TENT  & 71.10 & 0.0009 \\
           NOTE  & 42.59 & 0.0020 \\
           COTTA & 71.32 & 0.0190 \\
           TTAC (w/o queue) & 43.40 & 0.0021 \\
           PETAL & 71.14 & 0.0163 \\
           ROTTA & 30.50 & 0.0033 \\
        \midrule
           \bf TRIBE & \bf 19.53 & 0.0043 \\
        \bottomrule[1.2pt]
        \end{tabular}
    }
    \caption{The per-sample wall time (measured in seconds) on CIFAR10-C under GLI-TTA-F (IF=100) protocol.}
    \label{tab:wall_clock}
\end{table}

\subsection{The robustness of TRIBE under more fine-grained degrees of global class-imbalance under GLI-TTA-F protocol}

We evaluate TRIBE on more fine-grained degrees of global class-imbalance under GLI-TTA-F protocol and show the instance-level average classification error rates on Tab.~\ref{tab:Different_global_imbalanced_degree}. We choose the global imbalance factor as $1, 2, 4, 8, 10, 40, 80, 100$ and $200$. The results demonstrate the robustness and stability of our proposed \textbf{TRIBE} on different global class-imbalanced degrees, and the performance varies between $16.10\%$ and $20.98\%$.

\begin{table}[!ht]
    \centering
    \caption{More detailed results under different global class-imbalance factors on CIFAR10-C under GLI-TTA-F protocol. The instance-level average error rates are reported.}
    \resizebox{\linewidth}{!}{
        \begin{tabular}{c|cccccccccc}
        \toprule[1.2pt]
            I.F. & 1 & 2 & 4 & 8 & 10 & 20 & 40 & 80 & 100 & 200 \\
        \midrule
            \bf TRIBE & 16.10 & 18.97 & 20.15 & 20.96 & 20.98 & 20.63 & 20.01 & 19.37 & 19.53 & 19.16\\
        \bottomrule[1.2pt]
        \end{tabular}
    }
    \label{tab:Different_global_imbalanced_degree}
\end{table}

\subsection{The robustness of TRIBE under different degrees of local class-imbalance under practical TTA protocol}

We evaluate TRIBE on different degrees of local class-imbalance. The comparisons between RoTTA~\cite{yuan2023robust} and our proposed \textbf{TRIBE} on three datasets (i.e. CIFAR10-C, CIFAR100-C and ImageNet-C) are shown in Tab.~\ref{tab:Different_local_imbalanced_degree}. We make the following observations. First, our proposed \textbf{TRIBE} consistently outperforms RoTTA on all datasets at all degrees of local class-imbalance. Second, with $\sigma$ becoming smaller, the local class-imbalance is becoming more severe, but \textbf{TRIBE} is still able to maintain the performance on CIFAR10-C while RoTTA becomes substantially worse. This is probably attributed to the ability to welll balance between self-training loss with strong adaptive ability and anchored loss with stable regularized ability. Third, \textbf{TRIBE} has smaller performance fluctuations with the changing of local class-imbalanced degrees, suggesting the robustness of TRIBE due to more sophisticated Balanced BN layer and tri-net architecture.

\begin{table}[!ht]
    \centering
    \caption{Average classification errors of different local class-imbalanced degrees $\sigma$ on three datasets. non-i.i.d. indicates more correlated testing data stream which results in local class-imbalance. i.i.d. indicates the neighborhood samples are sampled from i.i.d. distribution.}
    \resizebox{\textwidth}{!}{
    \begin{tabular}{c|c|lllll}
    \toprule[1.2pt]
        \multirow{2}{*}{Dataset} & \multirow{2}{*}{Method} & \multicolumn{5}{c}{$i.i.d.\xrightarrow{\hspace*{5cm}} non-i.i.d.$} \\
        & & 10. & 1.0 & 0.1 & 0.01 & 0.001 \\
    \midrule
        \multirow{2}{*}{CIFAR10-C} & RoTTA & 20.45 & 21.15 & 25.20 & 28.69 & 29.11 \\
        & \method & \bf 19.91\dtplus{+0.54} & \bf 19.88\dtplus{+1.27} & \bf 16.14\dtplus{+9.06} & \bf 14.47\dtplus{+14.22} & \bf 12.39\dtplus{+16.72}\\
        % & \method~(Gamma: 0.1) & 19.75 & 19.79 & 16.24 & 14.47 & 13.02\\
    \midrule
        \multirow{2}{*}{CIFAR100-C} & RoTTA & 33.90 & 34.00 & 35.00 & 38.20 & 40.14\\
        & \method & \bf 32.69\dtplus{+1.21} & \bf 32.93\dtplus{+1.07} & \bf 33.26\dtplus{+1.74} & \bf 34.50\dtplus{+3.70} & \bf 36.02\dtplus{+4.12}\\
    \midrule
        \multirow{2}{*}{ImageNet-C} & RoTTA & 66.70 & 66.81 & 66.90 & 67.20 & 67.33\\
        & \method & \bf 61.84\dtplus{+4.86} & \bf 62.53\dtplus{+4.28} & \bf 62.32\dtplus{+4.58} & \bf 63.02\dtplus{+4.18} & \bf 63.33\dtplus{+4.00}\\
    \bottomrule[1.2pt]
    \end{tabular}
    }
    \label{tab:Different_local_imbalanced_degree}
\end{table}

\subsection{Sensitivity analysis of hyper-parameters used in TRIBE}

\noindent\textbf{Sensitiveness of the balancing parameter $\gamma$ in Balanced BN}: To evaluate the sensitiveness of $\gamma$, we conduct the experiments on GLI-TTA-F ($I.F.=100$) protocols on CIFAR10-C and CIFAR100-C. We show the performance from $\gamma=0.0$ to $\gamma=1.0$ in Tab.~\ref{tab:BN_gamma}. Specifically, $\gamma=1.0$ indicates the Balanced BN (class-wise statistics) degenerates to Robust BN (class-agnostic statistics). In this work, we choose $\gamma=0.0$ on CIFAR10-C and $\gamma=0.1$ on CIFAR100-C dataset. Additionally, we observe that, on CIFAR10-C dataset, when $\gamma=1.0$ the performance degrades severely (Inst. Avg.: $19.53\% \to 43.26\%$) and it demonstrates the class-agnostic batch norm is sensitive and vulnerable to test data stream with severe global and local class-imbalance. On CIFAR100-C dataset, the degradation is relatively mild (Inst. Avg.: $32.31\% \to 33.26\%$) and this is probably caused by the fact that the number of testing samples per class on CIFAR100-C is much smaller than that on CIFAR10-C test set (shown in Fig.~\ref{fig:ClassDistCurve_C10} and Fig.~\ref{fig:ClassDistCurve_C100}).% and the harm on CIFAR100-C from smaller number of local class-imbalanced samples is slighter than that on CIFAR10-C.

\begin{table*}[!ht]
    \centering
    \caption{The analysis about sensitivity to hyper-parameter $1-\gamma$. The classification error rates are reported on CIFAR10-C and CIFAR100-C datasets under GLI-TTA-F ($I.F.=100$) protocol.}
    \resizebox{\textwidth}{!}{
        \begin{tabular}{l|l|ccccccccccc}
        \toprule[1.2pt]
           Dataset & Error Rate & 1.0  & 0.9 & 0.8 & 0.7 & 0.6 & 0.5 & 0.4 & 0.3 & 0.2 & 0.1 & 0.0 (Robust BN) \\
        \midrule
           \multirow{2}{*}{CIFAR10-C} & Inst. Avg. & \bf 19.53 & 20.33 & 21.35 & 21.87 & 22.71 & 23.69 & 24.40 & 25.60 & 26.74 & 29.02 & 43.26\\
           & Cat. Avg. & 24.66 & \bf 24.19 & 24.51 & 24.36 & 24.33 & 24.82 & 24.63 & 25.23 & 25.51 & 25.99 & 31.49\\
        \midrule
            \multirow{2}{*}{CIFAR100-C} & Inst. Avg. & 32.98 & \bf 32.31 & 32.85 & 32.71 & 32.69 & 32.58 & 32.61 & 32.72 & 32.67 & 32.80 & 33.26 \\
            & Cat. Avg. & 35.17 & \bf 34.98 & 35.03 & 35.06 & 35.03 & 35.18 & 35.39 & 35.49 & 35.74 & 35.52 & 35.54\\
        \toprule[1.2pt]
        \end{tabular}
    }
    \label{tab:BN_gamma}
\end{table*}

\noindent\textbf{Sensitiveness to the thresholding parameter $H_0$ in Tri-Net Self-Training}: To evaluate the sensitiveness of $H_0$, we conduct the experiments on GLI-TTA-F ($I.F.=100$) protocols on CIFAR10-C and CIFAR100-C. We show the performance from $H_0=0.0$ to $H_0=0.4$ in Tab.~\ref{tab:Entropy_threshold}. In this work, we choose $H_0=0.05$ on CIFAR10-C and $H_0=0.20$ on CIFAR100-C for the experiments conducted on the manuscript. The $H_0$ is the hyper-parameter used in Tri-Net Self-Training module and aims to filter the high reliable pseudo labels for the stably adaptation process. This hyper-parameter is highly related to the accuracy on the original dataset (e.g. clean test set), yet we found the performance is continuous and robust from $H_0=0.05$ to $H_0=0.40$ on both CIFAR10-C and CIFAR100-C datasets.

\begin{table*}[!ht]
    \centering
    \caption{The sensitive analysis about hyper-parameter $H_0$. The classification error rates are reported on CIFAR10-C and CIFAR100-C datasets under GLI-TTA-F ($I.F.=100$) protocol.}
    % \resizebox{\textwidth}{!}{
        \begin{tabular}{l|l|cccccccc}
        \toprule[1.2pt]
           Dataset & Error Rate & 0.05 & 0.10 & 0.15 & 0.20 & 0.25 & 0.30 & 0.35 & 0.40 \\
        \midrule
           \multirow{2}{*}{CIFAR10-C} & Inst. Avg. & \bf 19.53 & 20.00 & 20.15 & 20.14 & 20.54 & 20.63 & 20.96 & 20.96\\
           & Cat. Avg. & 24.66 & \bf 24.50 & 24.56 & 24.50 & 24.79 & 24.57 & 24.57 & 24.50\\
        \midrule
            \multirow{2}{*}{CIFAR100-C} & Inst. Avg. & 33.35 & 32.60 & 32.53 & 32.31 & 32.36 & 32.42 & \bf 32.11 & 32.53\\
            & Cat. Avg. & 35.30 & 35.96 & 34.92 & 34.98 & 35.34 & 35.35 & \bf 34.79 & 35.02\\
        \toprule[1.2pt]
        \end{tabular}
    % }
    \label{tab:Entropy_threshold}
\end{table*}

\noindent\textbf{Sensitivity to $\lambda_{anc}$}: To evaluate the sensitiveness of $\lambda_{anc}$, we conduct the experiments on GLI-TTA-F ($I.F.=100$) protocols on CIFAR10-C and CIFAR100-C. We show the performance from $\lambda_{anc}=0.0$ to $\lambda_{anc}=2.0$ in Tab.~\ref{tab:Loss_scale}. $\lambda_{anc}$ is the hyper-parameter balancing $\mathcal{L}_{st}$ and $\mathcal{L}_{anc}$, where $\mathcal{L}_{st}$ provides the strong adaptation ability to the specific domain but it easily causes the accumulated confirmation bias and over-confident issue~\cite{arazo2020pseudo, su2023revisiting} and $\mathcal{L}_{anc}$ provides the regularization constraint to self-training. We choose $\lambda_{anc}=0.5$ on all experiments of the manuscript. We observe that without $\mathcal{L}_{anc}$ ($\lambda_{anc}=0.$) the performance degrades from $19.53\%$ to $36.58\%$ on CIFAR10-C and from $32.31\%$ to $37.21\%$ on CIFAR100-C which demonstrates the important role of the anchored loss $\mathcal{L}_{anc}$.

\begin{table*}[!ht]
    \centering
    \caption{The sensitive analysis about hyper-parameter $\lambda_{anc}$. The classification error rates are reported on CIFAR10-C and CIFAR100-C datasets under GLI-TTA-F ($I.F.=100$) protocol.}
    \begin{tabular}{c|c|ccccc}
    \toprule[1.2pt]
        Dataset & Error Rate & 0.0 & \bf 0.5 & 1.0 & 1.5 & 2.0 \\
    \midrule
        \multirow{2}{*}{CIFAR10-C} & Inst. Avg. & 36.58 &  \bf 19.53 & 21.94 & 23.23 & 24.32\\
        & Cat. Avg. & 65.88 & 24.66 & 24.52 &  24.43 & 24.69\\
    \midrule
        \multirow{2}{*}{CIFAR100-C} & Inst. Avg. & 37.21 &  \bf 32.31 & 33.93 & 34.95 & 35.53\\
        & Cat. Avg. & 44.83 & 34.98 &  34.92 & 35.18 & 35.17\\
    \bottomrule[1.2pt]
    \end{tabular}
    \label{tab:Loss_scale}
\end{table*}

\noindent\textbf{Sensitivity to $\eta$}: To evaluate the sensitivenss of $\eta$, we conduct the experiments on GLI-TTA-F ($I.F.=100$) protocols on CIFAR10-C and CIFAR100-C. We show the performance from $\eta=5e-5 \times K_c$ to $\eta=1e-2 \times K_c$ in Tab.~\ref{tab:Eta}.

\begin{table*}[!ht]
    \centering
    \caption{The sensitive analysis about hyper-parameter $\eta$. The classification error rates are reported on CIFAR10-C and CIFAR100-C datasets under GLI-TTA-F ($I.F.=100$) protocol.}
    \begin{tabular}{c|cccc}
    \toprule[1.2pt]
         Dataset & $5e-5 \times K_c$ & $5e-4 \times K_c$ & $5e-3 \times K_c$ & $1e-2 \times K_c$ \\
    \midrule
        CIFAR10-C & 34.04 &   \bf 19.53 &       22.73      &      26.04       \\
        CIFAR100-C & 45.57 &   \bf 32.31 &       32.66      &      32.85       \\
    \bottomrule[1.2pt]
    \end{tabular}
    \label{tab:Eta}
\end{table*}

% \subsection{Evaluation on the computation cost}
% \change{We evaluate the computation cost of our proposed TRIBE in this section. It's worthy note that, we only update the weight and bias parameters in batchnorm layers so that most of parameters across three models are shared, with only a small fraction of the remaining independent parameters. On the other hand, we have implemented Balanced BN as efficiently as possible using cpp in order to make the algorithm more efficient and more effective. Here, we provide the computation cost analysis of our TRIBE and several SOTA TTA methods in Tab.~\ref{tab:wall_clock}. We make the following observations. Our model undeniably got the best result and by a huge margin with others, and our model didn't take too much extra time compared to other SOTAs (only 1ms more than ROTTA), which we feel is acceptable for a TTA method.}

% \begin{table}[!h]
%     \centering
%     \begin{tabular}{c|c|c}
%     \toprule[1.2pt]
%        Method  &  Error Rate & Inference and Adaptation Time per Sample (second)\\
%     \midrule
%        TEST  & 41.71 & 0.0005 \\
%        BN    & 70.00 & 0.0005 \\
%        TENT  & 71.10 & 0.0009 \\
%        NOTE  & 42.59 & 0.0020 \\
%        COTTA & 71.32 & 0.0190 \\
%        TTAC (w/o queue) & 43.40 & 0.0021 \\
%        PETAL & 71.14 & 0.0163 \\
%        ROTTA & 30.50 & 0.0033 \\
%     \midrule
%        \bf TRIBE & \bf 19.53 & 0.0043 \\
%     \bottomrule[1.2pt]
%     \end{tabular}
%     \caption{The per-sample wall time (measured in seconds) on CIFAR10-C under GLI-TTA-F (IF=100) protocol.}
%     \label{tab:wall_clock}
% \end{table}

\section{Related Work}

\subsection{Test-time Strategies}

\begin{itemize}
\item \cite{wu2021online} proposed Online gradient descent and Follow The History algorithms to adapt model to handle the label shift appearing at test time. However, there is no domain shift considered between training and test set. There's no denying these are great methods but it would make the pseudo labels (gradient direction from minimizing entropy) untrustworthy when the domain is changed.

\item \cite{zhang2022self} adopted three expert models, which supervised by three different losses, to learn from the same long-tailed source data in order to fuse them to deal with unknown test class distributions. However, the difficulties in TTA are i) the source model is pretrained in advance and we can not utilize the source domain to again train multiple expert models to deal with some test distributions, ii) training in long tailed source data with GT labels is different from fine-tuning models in long tailed test data with pseudo labels, since self-training with pseudo labels would easily accumulate confirmation bias without the regularization from GT labels. 
\end{itemize}

In this paper, we mainly address the problem that given a pretrained source model (No special treatment for the test environment, generic model), we need to adapt it into a more realistic test scenario which fuses locally and globally class imbalanced testing data and the test environment (corruptions) is time-varying. Unlike reweighting strategy used into long-tailed tasks, e.g. Balanced Softmax, we propose the Balanced BatchNorm to alleviate the estimation bias of statistics maintained in batchnorm layers, and propose Anchor Network with Anchored Loss to alleviate the confirmation bias of learnable parameters updated by self-training loss.

\section{Derivation of $\mu_g$ and $\sigma_g^2$ in Balanced BN}

Balanced BN aims to obtain the class-balanced global distribution according to testing samples consisting of different numbers of samples from different category-wise distributions. Here, we provide the derivation of Balanced BN motivated by imbalanced testing data stream. We first assume that the testing samples $\{\vect{x}_i\}_{i=1\cdots N}$ consists of $K_c$ categories of $\{\vect{x}_{k,i}\}_{i=1\cdots N_k}$, where $N_k$ denotes the number of samples belonging to class $k$. We denote the $k^{th}$ categorical distribution as $\{\mu_k, \sigma_k^2\}$. The class-agnostic global distribution $\{\mu_g, \sigma_g^2\}$ in a regular normalization layer (e.g. BN~\cite{ioffe2015batch}) can be obtained as Eq.~\ref{eq:mu_g} and Eq.~\ref{eq:sigma_g}. 

\begin{equation}\label{eq:mu_g}
    \mu_g = \frac{\sum_{i=1}^N \vect{x}_i}{N} = \frac{\sum_{k=1}^{K_c} \sum_{i=1}^{N_k} \vect{x}_{k,i}}{\sum_{k=1}^{K_c} N_k} = \frac{\sum_{k=1}^{K_c} N_k \mu_k}{\sum_{k=1}^{K_c} N_k},
\end{equation}

\begin{equation}\label{eq:sigma_g}
\begin{aligned}
\sigma^2_g & = \frac{\sum_{i=1}^N (\vect{x}_i - \mu_g)^2}{N} = \frac{\sum_{k=1}^{K_c} \sum_{i=1}^{N_k} (\vect{x}_{k,i} - \mu_g)^2}{\sum_{k=1}^{K_c} N_k} \\
& = \frac{1}{\sum_{k=1}^{K_c} N_k} \sum_{k=1}^{K_c} \sum_{i=1}^{N_k} (\vect{x}_{k,i} - \mu_k + \mu_k - \mu_g)^2 \\
& = \frac{1}{\sum_{k=1}^{K_c} N_k} \sum_{k=1}^{K_c} \sum_{i=1}^{N_k} (\vect{x}_{k,i} - \mu_k)^2 + (\mu_k - \mu_g)^2 \\ 
& + 2(\vect{x}_{k,i} - \mu_k)(\mu_k - \mu_g) \\
& = \frac{\sum_{k=1}^{K_c} N_k \left[ \sigma^2_k + (\mu_k - \mu_g)^2 \right]}{\sum_{k=1}^{K_c} N_k}.
\end{aligned}
\end{equation}

We can observe that the $\mu_g$ and $\sigma_g^2$ in regular BN tend to approximate to the distribution of the most majority class when the testing samples are severely class-imbalanced either locally or globally $\mu_g \approx \mu_{k'}, \sigma^2_g \approx \sigma^2_{k'}\quad s.t.\quad N_{k'} \approx N$, which would introduce a large bias to the model pre-trained on class-balanced source domain.

To mitigate the impact from the locally or globally class-imbalanced testing samples, we re-weight the category-wise distributions to obtain the class-balanced global distribution (denoted as $\{\mu_{ba}, \sigma_{ba}^2\}$) as Eq.~\ref{eq:update_mu} and Eq.~\ref{eq:update_sigma}.

\begin{equation}\label{eq:update_mu}
    \mu_{ba} = \frac{1}{K_c} \sum_{k=1}^{K_c} \frac{\sum_{i=1}^{N_k} \vect{x}_{k,i}}{N_k} = \frac{1}{K_c} \sum_{k=1}^{K_c} \mu_k
\end{equation}

\begin{equation}
\begin{aligned}
    \sigma^2_{ba} & = \frac{1}{K_c} \sum_{k=1}^{K_c} \frac{\sum_{i=1}^{N_k} (\vect{x}_{k,i} - \mu_{ba})^2 }{N_k} \\
    & = \frac{1}{K_c} \sum_{k=1}^{K_c} \frac{\sum_{i=1}^{N_k} (\vect{x}_{k, i} - \mu_k + \mu_k - \mu_{ba})^2 }{N_k} \\
    & = \frac{1}{K_c\cdot N_k} \sum_{k=1}^{K_c} \sum_{i=1}^{N_k} (\vect{x}_{k, i} - \mu_k)^2 + (\mu_k - \mu_{ba})^2 \\ 
    & + 2(\vect{x}_{k, i} - \mu_k)(\mu_k - \mu_{ba}) \\
    & = \frac{1}{K_c} \sum_{k=1}^{K_c} \left[ \sigma_k^2 + (\mu_k - \mu_{ba})^2 \right]
\end{aligned}\label{eq:update_sigma}
\end{equation}

The formulas of $\mu_{ba}$ and $\sigma_{ba}$ above demonstrate the ability of estimating the global class-balanced distribution using class-imbalanced testing samples due to the re-weighting strategy. We utilize the pseudo labels of testing samples to categorize them into $K_c$ categories, and the pseudo labels are often predicted at the inference step of TTA.

% \section{Theoretical proof of Balanced BN adapting class-imbalanced data}

\section{Algorithm and implementation of Balanced BN}

We provide an overview of the algorithm (Alg.~\ref{alg:BalancedBN}) of Balanced BN in this section. An efficient implementation of Balanced BN using \textit{Cpp} is also attached in the supplementary material.

\begin{algorithm}[!ht]
\SetKwInOut{Input}{input}
\SetKwInOut{Output}{output}
\Input{{Input feature map $\mathbf{F}\in \mathbb{R}^{B\times C\times H\times W}$.
Corresponding pseudo labels $\hat{\mathbf{Y}}\in\mathbb{R}^{B\times K_c}$.
}}

\Output{Normalized feature map $\tilde{\mathbf{F}}$.}

\If{training}
{
    update $\{\mu_k\}_{k=1\cdots K_c}$ and $\{\sigma_k^2\}_{k=1\cdots K_c}$ according to Eq.~3 of the manuscript.
}

calculate $\mu_{ba}$ and $\sigma^2_{ba}$ according to Eq.~\ref{eq:update_mu} and Eq.~\ref{eq:update_sigma}.

$\tilde{\mathbf{F}}=\mathbf{\alpha} \frac{\mathbf{F} - \mu_{ba}}{\sqrt{\sigma^2_{ba}+\epsilon}} + \mathbf{\beta}$,

where $\mathbf{\alpha}$ and $\mathbf{\beta}$ here indicate the affine parameters of standard batchnorm module.

\Return $\tilde{\mathbf{F}}$

\caption{\small{The algorithm of Balanced BN}\label{alg:BalancedBN}}
\end{algorithm}

\section{Data augmentations used in TRIBE}

Data augmentation plays an important role in self-training. In this section, we provide more details of the data augmentation strategy adopted for the teacher model in TRIBE. The augmentation strategy is similar to the ones adopted in RoTTA~\cite{yuan2023robust}. The data augmentations are including Clip, ColorJitter, Pad, RandomAffine and etc. The specific details can be found in the code provided into the supplementary materials.

\section{Detail results under different GLI-TTA protocols}

Finally, we present more detailed results for each dataset under the proposed TTA protocols.

\begin{table*}[t]
    \centering
    \caption{Average classification error on CIFAR10-C while continually adapting to different corruptions at the highest severity 5 with globally and locally class-imbalanced test stream and fixed global class distribution (GLI-TTA-F). $I.F.$ is the Imbalanced Factor of Global Class-Imbalance.}
    \label{tab:DLI_CIFAR10}
    \resizebox{\textwidth}{!}{
    {
    \begin{tabular}{l|l|ccccccccccccccc|cc}
        \toprule[1.2pt]
        \multicolumn{2}{c}{Time} & \multicolumn{15}{l|}{$t\xrightarrow{\hspace*{18.5cm}}$}& \\ \hline
        $I.F.$ & Method & \rotatebox[origin=c]{45}{motion} & \rotatebox[origin=c]{45}{snow} & \rotatebox[origin=c]{45}{fog} & \rotatebox[origin=c]{45}{shot} & \rotatebox[origin=c]{45}{defocus} & \rotatebox[origin=c]{45}{contrast} & \rotatebox[origin=c]{45}{zoom} & \rotatebox[origin=c]{45}{brightness} & \rotatebox[origin=c]{45}{frost} & \rotatebox[origin=c]{45}{elastic} & \rotatebox[origin=c]{45}{glass} & \rotatebox[origin=c]{45}{gaussian} & \rotatebox[origin=c]{45}{pixelate} & \rotatebox[origin=c]{45}{jpeg} & \rotatebox[origin=c]{45}{impulse}
        & Inst.Avg. & Cat.Avg.\\ 
        
        \midrule
        \multirow{9}{*}{10} & TEST & 29.48 & 25.64 & 26.30 & 65.72 & 42.92 & 44.64 & 34.94 & 9.30 & 43.61 & 25.81 & 53.97 & 72.01 & 63.00 & 33.18 & 69.27 & 42.65 & 43.79\\

        & BN~\cite{BN_Stat} & 66.04 & 70.30 & 68.90 & 72.38 & 69.17 & 68.83 & 68.54 & 65.60 & 68.24 & 74.49 & 75.44 & 73.14 & 69.78 & 74.80 & 75.91 & 70.77 & 66.77 \\
        
        & PL~\cite{PL} & 66.58 & 70.64 & 69.49 & 73.53 & 71.52 & 70.64 & 70.15 & 68.17 & 69.76 & 75.00 & 78.04 & 74.85 & 73.29 & 77.74 & 76.98 & 72.43 & 70.59\\
        
        & TENT~\cite{tent_wang2020} & 66.58 & 70.74 & 70.74 & 75.39 & 73.16 & 74.76 & 75.34 & 75.32 & 78.70 & 82.37 & 84.75 & 84.72 & 83.67 & 87.90 & 88.10 & 78.15 & 74.90\\

        & LAME~\cite{niid_boudiaf2022parameter} & 23.70 & 19.81 & 20.67 & 65.48 & 37.73 & 41.53 & 29.16 & \bf 5.73 & 39.45 & \bf 17.70 & 50.32 & 71.96 & 61.51 & \bf 24.85 & 67.16 & 38.45 & 40.07\\

        & NOTE~\cite{note} & 19.76 & 23.82 & 21.72 & 46.25 & 26.57 & 15.11 & 27.06 & 20.98 & 39.47 & 46.67 & 59.72 & 51.00 & 53.28 & 47.06 & 53.33 & 36.79 & 30.22\\

        & TTAC~\cite{su2022revisiting} & 31.78 & 28.11 & 21.94 & 66.45 & 27.11 & 20.08 & 19.10 & 10.11 & 25.73 & 23.73 & 42.46 & 34.35 & 35.92 & 27.89 & 53.21 & 31.20 & 29.11\\

        & RoTTA~\cite{yuan2023robust} & 20.13 & 22.72 & 16.41 & 37.46 & 20.32 & 17.04 & 18.58 & 11.24 & 25.37 & 31.22 & 45.64 & 36.04 & 30.97 & 31.29 & 46.56 & 27.41 & 26.31 \\

        \cmidrule{2-19}
        & \method & \bf 17.43 & \bf 18.12 & \bf 11.85 & \bf 25.66 & \bf 17.53 & \bf 12.83 & \bf 13.81 & 9.13 & \bf 16.45 & 22.26 & \bf 34.97 & \bf 24.90 & \bf 23.70 & 25.98 & \bf 40.06 & \bf 20.98\dtplus{+6.43} & \bf 22.49\dtplus{+3.82} \\

        \midrule
        \multirow{9}{*}{100} & TEST & 25.71 & 26.39 & 27.68 & 67.88 & 41.32 & 41.40 & 31.92 & 7.14 & 46.21 & 24.21 & 49.11 & 74.21 & 61.50 & 33.45 & 67.55 & 41.71 & 43.63\\

        & BN~\cite{BN_Stat} & 64.77 & 69.09 & 68.16 & 72.48 & 68.89 & 67.64 & 63.96 & 62.47 & 65.78 & 72.92 & 75.50 & 74.86 & 71.51 & 74.66 & 77.28 & 70.00 & 50.72\\

        & PL~\cite{PL} & 64.77 & 69.29 & 68.24 & 73.77 & 70.54 & 67.35 & 65.74 & 64.45 & 68.48 & 72.40 & 75.95 & 73.77 & 69.33 & 72.32 & 74.90 & 70.09 & 55.29\\
        
        & TENT~\cite{tent_wang2020} & 64.81 & 68.60 & 67.19 & 72.28 & 69.05 & 67.68 & 65.54 & 64.81 & 67.88 & 73.33 & 77.00 & 77.16 & 73.85 & 78.93 & 78.41 & 71.10 & 58.59\\

        & LAME~\cite{niid_boudiaf2022parameter} & 20.10 & \bf 19.65 & 21.95 & 67.84 & 36.64 & 37.77 & 25.14 & \bf 3.75 & 42.49 & \bf 16.67 & 44.75 & 74.29 & 60.17 & 25.34 & 65.70 & 37.48 & 41.80\\

        & NOTE~\cite{note} & 21.31 & 27.28 & 28.13 & 52.18 & 31.36 & 17.35 & 30.67 & 21.67 & 46.13 & 54.24 & 66.95 & 62.31 & 55.73 & 57.79 & 65.78 & 42.59 & 30.75\\

        & TTAC~\cite{su2022revisiting} & 40.96 & 48.59 & 31.56 & 77.60 & 47.18 & 31.80 & 24.05 & 13.44 & 44.75 & 31.03 & 46.53 & 58.64 & 42.62 & 33.41 & 78.81 & 43.40 & 37.37\\

        & RoTTA~\cite{yuan2023robust} & 22.40 & 31.19 & 21.71 & 44.51 & 27.56 & 14.89 & 23.33 & 10.13 & 26.96 & 32.69 & 44.87 & 41.53 & 34.22 & 31.76 & 49.68 & 30.50 & 29.08\\

        \cmidrule{2-19}
        & \method & \bf 17.43 & 20.34 & \bf 12.11 & \bf 28.17 & \bf 17.59 & \bf 11.38 & \bf 11.50 & 6.82 & \bf 15.50 & 18.85 & \bf 29.54 & \bf 23.69 & \bf 24.78 & \bf 24.37 & \bf 30.91 & \bf 19.53\dtplus{+10.97} & \bf 24.66\dtplus{+4.42}\\

        \midrule
        \multirow{9}{*}{200} & TEST & 25.16 & 26.32 & 28.37 & 69.35 & 40.88 & 40.44 & 31.72 & 6.75 & 47.45 & 24.08 & 48.17 & 75.29 & 60.41 & 33.07 & 67.92 & 41.69 & 43.47\\

        & BN~\cite{BN_Stat} & 64.88 & 70.42 & 68.63 & 73.68 & 70.02 & 65.24 & 65.19 & 64.66 & 67.16 & 71.00 & 74.93 & 74.62 & 69.57 & 73.55 & 78.33 & 70.13 & 47.34\\

        & PL~\cite{PL} & 65.33 & 70.78 & 67.92 & 73.99 & 69.71 & 65.10 & 64.97 & 64.30 & 66.89 & 71.63 & 76.01 & 75.34 & 71.18 & 75.60 & 76.99 & 70.38 & 49.86\\
        
        & TENT~\cite{tent_wang2020} & 65.19 & 71.13 & 68.54 & 73.73 & 70.42 & 65.95 & 65.68 & 64.39 & 66.67 & 69.71 & 73.41 & 70.69 & 67.38 & 70.42 & 73.86 & 69.15 & 53.37\\

        & LAME~\cite{niid_boudiaf2022parameter} &  19.21 & 19.62 & 22.79 & 70.78 & 34.99 & 37.04 & 25.83 & \bf 3.84 & 43.66 & \bf 16.53 & 42.85 & 75.69 & 58.94 & 25.60 & 65.46 & 37.52 & 42.59\\

        & NOTE~\cite{note} & 21.98 & 29.71 & 33.51 & 58.45 & 35.61 & 21.31 & 36.19 & 24.22 & 48.93 & 55.54 & 67.29 & 65.91 & 56.66 & 58.71 & 67.69 & 45.45 & 31.17\\

        & TTAC~\cite{su2022revisiting} & 43.66 & 55.90 & 36.06 & 75.69 & 49.29 & 35.17 & 23.46 & 13.45 & 50.27 & 33.91 & 47.14 & 71.31 & 46.60 & 36.33 & 75.83 & 46.27 & 38.75\\

        & RoTTA~\cite{yuan2023robust} & 23.95 & 32.84 & 23.28 & 48.08 & 29.13 & 16.76 & 25.11 & 10.19 & 29.94 & 36.95 & 46.65 & 45.80 & 34.94 & 33.69 & 49.51 & 32.45 & 30.04 \\

        \cmidrule{2-19}
        & \method & \bf 17.16 & \bf 18.99 & \bf 10.46 & \bf 31.77 & \bf 18.01 & \bf 10.72 & \bf 11.39 & 7.10 & \bf 14.34 & 18.72 & \bf 29.31 & \bf 21.49 & \bf 23.19 & \bf 23.68 & \bf 31.10 & \bf 19.16\dtplus{+13.29} & \bf 24.00\dtplus{6.04} \\
        
    \bottomrule[1.2pt]
    \end{tabular}
    }
    }
\end{table*}

\begin{table*}[t]
    \centering
    \caption{Average classification error on CIFAR100-C while continually adapting to different corruptions at the highest severity 5 with globally and locally class-imbalanced test stream and fixed global class distribution (GLI-TTA-F). $I.F.$ is the Imbalanced Factor of Global Class-Imbalance.}
    \label{tab:DLI_CIFAR100}
    \resizebox{\textwidth}{!}{
    {
    \begin{tabular}{l|l|ccccccccccccccc|cc}
        \toprule[1.2pt]
        \multicolumn{2}{c}{Time} & \multicolumn{15}{l|}{$t\xrightarrow{\hspace*{18.5cm}}$}& \\ \hline
        I.F. & Method & \rotatebox[origin=c]{45}{motion} & \rotatebox[origin=c]{45}{snow} & \rotatebox[origin=c]{45}{fog} & \rotatebox[origin=c]{45}{shot} & \rotatebox[origin=c]{45}{defocus} & \rotatebox[origin=c]{45}{contrast} & \rotatebox[origin=c]{45}{zoom} & \rotatebox[origin=c]{45}{brightness} & \rotatebox[origin=c]{45}{frost} & \rotatebox[origin=c]{45}{elastic} & \rotatebox[origin=c]{45}{glass} & \rotatebox[origin=c]{45}{gaussian} & \rotatebox[origin=c]{45}{pixelate} & \rotatebox[origin=c]{45}{jpeg} & \rotatebox[origin=c]{45}{impulse}
        & Inst.Avg. & Cat.Avg.\\ 
        
        \midrule
        \multirow{9}{*}{10} & TEST & 30.06 & 40.25 & 51.60 & 69.32 & 29.95 & 55.60 & 28.97 & 29.57 & 47.21 & 36.89 & 54.10 & 73.68 & 76.70 & 41.23 & 39.27 & 46.96 & 46.52\\

        & BN~\cite{BN_Stat} & 39.99 & 46.44 & 52.48 & 49.95 & 39.81 & 41.10 & 39.29 & 37.93 & 46.16 & 46.67 & 51.01 & 52.66 & 44.76 & 50.75 & 51.75 & 46.05 & 42.29 \\
        
        & PL~\cite{PL} & 40.12 & 46.47 & 55.01 & 57.17 & 54.15 & 60.71 & 62.82 & 64.01 & 71.18 & 77.63 & 83.67 & 85.17 & 85.84 & 90.61 & 93.06 & 68.51 & 69.71\\
        
        & TENT~\cite{tent_wang2020} & 39.16 & 45.28 & 53.82 & 61.76 & 63.39 & 76.37 & 78.25 & 83.38 & 89.50 & 91.49 & 93.52 & 92.96 & 93.94 & 95.23 & 95.18 & 76.88 & 79.08\\

        & LAME~\cite{niid_boudiaf2022parameter} & \bf 26.08 & 34.80 & 48.35 & 68.99 & \bf 24.59 & 54.77 & \bf 23.66 & \bf 24.61 & 43.70 & \bf 31.09 & 50.77 & 73.81 & 78.22 & 36.56 & 34.93 & 43.66 & 44.88\\

        & NOTE~\cite{note} & 44.94 & 51.50 & 58.82 & 62.95 & 47.88 & 40.66 & 49.23 & 47.50 & 55.57 & 60.50 & 67.44 & 66.77 & 63.98 & 70.10 & 77.84 & 57.71 & 58.86\\

        & TTAC~\cite{su2022revisiting} & 28.87 & 41.28 & 46.65 & 56.01 & 33.69 & 44.53 & 31.76 & 29.44 & 40.09 & 37.33 & 43.42 & 50.88 & 46.08 & 42.23 & \bf 34.91 & 40.48 & 38.28\\

        & RoTTA~\cite{yuan2023robust} & 33.88 & 42.00 & 47.01 & 49.74 & 35.37 & 46.16 & 35.73 & 27.09 & 39.58 & 39.99 & 41.59 & 42.05 & 37.15 & 38.98 & 43.73 & 40.00 & 39.03 \\

        \cmidrule{2-19}
        & \method & 28.35 & \bf 33.13 & \bf 38.83 & \bf 39.11 & 28.33 & \bf 27.84 & 26.68 & 24.64 & \bf 31.22 & 33.23 & \bf 37.44 & \bf 39.65 & \bf 33.26 & \bf 36.38 & 38.42 & \bf 33.10\dtplus{+6.90} & \bf 34.31\dtplus{+3.97} \\

        \midrule
        \multirow{9}{*}{100} & TEST & 29.80 & 42.05 & 50.96 & 69.45 & 31.39 & 55.84 & 29.28 & 30.17 & 49.18 & 37.45 & 54.20 & 75.04 & 77.66 & 41.11 & 39.37 & 47.53 & 45.91\\

        & BN~\cite{BN_Stat} & 40.69 & 47.30 & 54.76 & 51.20 & 40.22 & 42.05 & 40.22 & 39.93 & 47.49 & 47.96 & 51.76 & 51.10 & 45.28 & 53.26 & 51.95 & 47.01 & 40.01\\

        & PL~\cite{PL} & 40.26 & 45.75 & 50.82 & 49.32 & 44.39 & 47.16 & 44.91 & 46.08 & 52.28 & 54.81 & 60.72 & 64.38 & 62.08 & 68.61 & 70.30 & 53.46 & 57.26\\
        
        & TENT~\cite{tent_wang2020} & 38.90 & 42.23 & 45.75 & 46.88 & 42.23 & 48.57 & 46.97 & 51.15 & 59.27 & 63.02 & 67.95 & 72.08 & 72.13 & 76.40 & 77.29 & 56.72 & 65.96\\

        & LAME~\cite{niid_boudiaf2022parameter} & \bf 25.53 & 36.37 & 46.88 & 68.84 & \bf 25.86 & 54.95 & \bf 23.98 & 24.50 & 46.27 & 31.53 & 50.59 & 75.74 & 79.02 & 35.95 & 36.32 & 44.15 & 46.64\\

        & NOTE~\cite{note} & 44.53 & 50.40 & 58.94 & 65.27 & 46.93 & 38.90 & 45.57 & 43.36 & 50.16 & 54.95 & 62.04 & 64.34 & 58.61 & 63.12 & 69.50 & 54.44 & 57.10\\

        & TTAC~\cite{su2022revisiting} & 31.11 & 52.28 & 52.37 & 61.10 & 36.88 & 53.73 & 32.75 & 33.18 & 49.23 & 43.97 & 45.71 & 60.63 & 66.64 & 54.34 & 43.74 & 47.84 & 41.47\\
        
        & RoTTA~\cite{yuan2023robust} & 35.05 & 46.27 & 51.10 & 57.72 & 42.05 & 40.36 & 45.89 & 28.63 & 49.27 & 50.45 & 46.03 & 50.45 & 48.33 & 46.93 & 46.69 & 45.68 & 42.04\\

        \cmidrule{2-19}
        & \method & 27.08 & \bf 34.02 & \bf 38.06 & \bf 41.20 & 29.94 & \bf 28.25 & 26.70 & \bf 23.51 & \bf 30.78 & \bf 31.44 & \bf 34.58 & \bf 37.45 & \bf 33.18 & \bf 34.73 & \bf 33.79 & \bf 32.31\dtplus{+11.84} & \bf 34.98\dtplus{+5.03}\\

        \midrule
        \multirow{9}{*}{200} & TEST & 29.86 & 41.30 & 50.54 & 69.71 & 30.99 & 56.39 & 29.22 & 30.72 & 49.68 & 37.92 & 54.51 & 74.87 & 77.82 & 41.14 & 39.26 & 47.59 & 39.94\\

        & BN~\cite{BN_Stat} & 41.46 & 47.48 & 53.76 & 52.31 & 40.33 & 43.02 & 41.41 & 40.33 & 47.42 & 47.69 & 52.09 & 51.02 & 44.58 & 54.78 & 53.01 & 47.38 & 35.26\\

        & PL~\cite{PL} & 41.35 & 43.72 & 47.80 & 48.01 & 39.69 & 44.36 & 41.62 & 39.96 & 48.07 & 50.27 & 55.96 & 57.30 & 54.51 & 64.45 & 64.07 & 49.41 & 49.26\\
        
        & TENT~\cite{tent_wang2020} & 39.42 & 42.00 & 45.76 & 44.58 & 39.53 & 43.02 & 40.01 & 42.32 & 49.14 & 54.19 & 59.72 & 60.96 & 58.11 & 67.51 & 70.46 & 50.45 & 58.45\\

        & LAME~\cite{niid_boudiaf2022parameter} & \bf 25.46 & 34.91 & 46.46 & 68.74 & \bf 25.46 & 54.46 & \bf 23.36 & 23.90 & 45.38 & \bf 31.69 & 51.07 & 75.56 & 79.86 & 35.39 & 35.45 & 43.81 & 40.33 \\

        & NOTE~\cite{note} & 44.95 & 50.27 & 59.67 & 63.48 & 46.19 & 37.65 & 45.44 & 43.72 & 50.48 & 54.83 & 61.82 & 61.49 & 57.84 & 61.06 & 67.19 & 53.74 & 52.48\\
        
        & TTAC~\cite{su2022revisiting} & 32.44 & 54.19 & 55.26 & 62.62 & 37.00 & 55.21 & 33.57 & 34.00 & 51.93 & 46.19 & 47.05 & 61.60 & 70.89 & 56.02 & 48.71 & 49.78 & 38.00\\

        & RoTTA~\cite{yuan2023robust} & 34.21 & 46.13 & 51.72 & 58.59 & 42.21 & 41.62 & 46.29 & 28.95 & 49.73 & 57.47 & 47.80 & 51.50 & 49.46 & 48.98 & 47.05 & 46.78 & 37.93 \\
        
        \cmidrule{2-19}
        & \method & 27.34 & \bf 32.17 & \bf 37.38 & \bf 40.01 & 30.50 & \bf 29.38 & 27.28 & \bf 23.58 & \bf 30.45 & 32.55 & \bf 34.91 & \bf 36.41 & \bf 35.82 & \bf 33.73 & \bf 32.81 & \bf 32.29\dtplus{+11.52} & \bf 31.54\dtplus{+3.72}\\
        
    \bottomrule[1.2pt]
    \end{tabular}
    }
    }
\end{table*}

\begin{table*}[t]
    \centering
    \caption{Average classification error on CIFAR10-C while continually adapting to different corruptions at the highest severity 5 with globally and locally class-imbalanced test stream and varying global class distribution (GLI-TTA-V). $I.F.$ is the Imbalanced Factor of Global Class-Imbalance.}
    \label{tab:DLI_VARY_CIFAR10}
    \resizebox{\textwidth}{!}{
    {
    \begin{tabular}{l|l|ccccccccccccccc|cc}
        \toprule[1.2pt]
        \multicolumn{2}{c}{Time} & \multicolumn{15}{l|}{$t\xrightarrow{\hspace*{18.5cm}}$}& \\ \hline
        I.F. & Method & \rotatebox[origin=c]{45}{motion} & \rotatebox[origin=c]{45}{snow} & \rotatebox[origin=c]{45}{fog} & \rotatebox[origin=c]{45}{shot} & \rotatebox[origin=c]{45}{defocus} & \rotatebox[origin=c]{45}{contrast} & \rotatebox[origin=c]{45}{zoom} & \rotatebox[origin=c]{45}{brightness} & \rotatebox[origin=c]{45}{frost} & \rotatebox[origin=c]{45}{elastic} & \rotatebox[origin=c]{45}{glass} & \rotatebox[origin=c]{45}{gaussian} & \rotatebox[origin=c]{45}{pixelate} & \rotatebox[origin=c]{45}{jpeg} & \rotatebox[origin=c]{45}{impulse}
        & Inst.Avg. & Cat.Avg.\\ 
        
        \midrule
        \multirow{9}{*}{10} & TEST & 34.33 & 24.66 & 24.44 & 63.42 & 40.23 & 41.82 & 39.59 & 10.09 & 37.61 & 24.12 & 51.18 & 81.32 & 56.22 & 30.63 & 69.61 & 41.95 & 43.65\\

        & BN~\cite{BN_Stat} & 66.45 & 71.35 & 69.56 & 72.82 & 68.41 & 67.19 & 70.89 & 64.10 & 69.05 & 74.80 & 76.52 & 72.48 & 73.16 & 75.86 & 77.82 & 71.36 & 67.70 \\
        
        & PL~\cite{PL} & 66.75 & 72.60 & 72.48 & 73.46 & 71.99 & 69.47 & 72.92 & 72.38 & 74.93 & 79.29 & 80.51 & 75.20 & 74.34 & 81.83 & 83.03 & 74.74 & 72.12\\
        
        & TENT~\cite{tent_wang2020} & 66.48 & 73.29 & 71.89 & 76.59 & 72.62 & 72.65 & 76.59 & 72.58 & 82.81 & 82.39 & 86.66 & 79.75 & 80.39 & 78.18 & 92.51 & 77.69 & 74.23\\

        & LAME~\cite{niid_boudiaf2022parameter} & 27.86 & \bf 18.66 & 18.95 & 63.76 & 35.58 & 38.86 & 34.72 & \bf 5.93 & 33.03 & \bf 17.07 & 47.33 & 82.13 & 55.24 & \bf 23.51 & 67.63 & 38.02 & 40.15\\

        & NOTE~\cite{note} & 17.38 & 22.97 & 17.34 & 37.76 & 24.83 & \bf 9.21 & 23.02 & 13.47 & 25.59 & 32.47 & 48.95 & 40.25 & 44.78 & 39.01 & 45.74 & 29.52 & 29.23\\

        & TTAC~\cite{su2022revisiting} & 18.93 & 25.76 & 16.80 & 62.41 & 29.31 & 17.90 & 17.12 & 11.83 & 23.16 & 21.18 & 43.19 & 42.41 & 58.57 & 31.61 & 63.61 & 32.25 & 32.12\\

        & RoTTA~\cite{yuan2023robust} & 14.79 & 27.50 & 16.23 & 37.10 & 31.86 & 15.50 & 16.38 & 12.49 & 23.16 & 32.05 & 43.41 & 32.84 & 33.77 & 30.66 & 46.35 & 27.61 & 26.35\\
        
        \cmidrule{2-19}
        & \method & \bf 12.78 & 20.20 & \bf 13.10 & \bf 25.32 & \bf 19.64 & 12.98 & \bf 13.57 & 9.04 & \bf 16.01 & 20.76 & \bf 33.25 & \bf 22.58 & \bf 31.51 & 26.13 & \bf 36.88 & \bf 20.92\dtplus{+6.69} & \bf 22.40\dtplus{+3.95}\\

        \midrule
        \multirow{9}{*}{100} & TEST & 32.32 & 23.57 & 24.50 & 64.49 & 34.02 & 37.57 & 40.11 & 9.52 & 32.04 & 23.20 & 47.05 & 85.84 & 58.88 & 31.64 & 66.38 & 40.74 & 43.83\\

        & BN~\cite{BN_Stat} & 67.35 & 70.54 & 69.49 & 70.58 & 66.42 & 66.95 & 71.03 & 61.86 & 64.97 & 71.87 & 75.46 & 71.87 & 74.17 & 74.86 & 77.80 & 70.35 & 53.07\\

        & PL~\cite{PL} & 67.51 & 70.42 & 70.22 & 72.52 & 65.70 & 69.45 & 71.99 & 63.24 & 69.33 & 71.55 & 82.85 & 78.53 & 82.16 & 72.48 & 87.49 & 73.03 & 57.53\\
        
        & TENT~\cite{tent_wang2020} & 68.28 & 70.90 & 70.98 & 71.19 & 67.96 & 68.60 & 70.34 & 66.06 & 74.29 & 72.96 & 81.48 & 80.23 & 73.61 & 68.40 & 89.59 & 72.99 & 58.65\\

        & LAME~\cite{niid_boudiaf2022parameter} & 25.67 & \bf 16.91 & 19.01 & 65.21 & 29.18 & 34.38 & 35.59 & \bf 5.33 & 26.84 & \bf 16.18 & 41.97 & 86.84 & 57.71 & \bf 24.33 & 62.55 & 36.51 & 42.16\\

        & NOTE~\cite{note} & 19.25 & 23.65 & 17.76 & 41.36 & 26.19 & \bf 7.99 & 27.48 & 13.12 & 25.30 & 24.90 & 46.81 & 39.95 & 48.47 & 41.85 & 46.17 & 30.02 & 29.88\\

        & TTAC~\cite{su2022revisiting} & 15.09 & 36.76 & 17.92 & 75.54 & 42.37 & 18.72 & 18.81 & 14.08 & 29.62 & 18.64 & 42.49 & 52.78 & 60.41 & 33.94 & 75.38 & 36.84 & 37.13\\

        & RoTTA~\cite{yuan2023robust} & 17.23 & 34.06 & 18.36 & 46.13 & 50.44 & 13.64 & 19.17 & 12.67 & 24.41 & 34.50 & 45.96 & 38.98 & \bf 41.44 & 34.71 & 50.65 & 32.16 & 29.32\\

        \cmidrule{2-19}
        & \method & \bf 13.03 & 23.49 & \bf 11.46 & \bf 29.26 & \bf 25.06 & 13.52 & \bf 13.60 & 7.75 & \bf 14.53 & 21.55 & \bf 32.20 & \bf 28.45 & 42.21 & 24.70 & \bf 35.84 & \bf 22.44\dtplus{+7.58} & \bf 25.50\dtplus{+3.82}\\

        \midrule
        \multirow{9}{*}{200} & TEST & 31.64 & 23.19 & 24.93 & 64.70 & 32.48 & 36.10 & 40.71 & 9.70 & 30.88 & 22.83 & 45.89 & 86.68 & 60.77 & 32.13 & 65.28 & 40.53 & 43.77\\

        & BN~\cite{BN_Stat} & 67.52 & 71.13 & 68.28 & 72.43 & 69.53 & 67.61 & 71.27 & 64.79 & 66.18 & 72.30 & 75.16 & 72.21 & 73.15 & 73.73 & 77.93 & 70.88 & 50.67 \\

        & PL~\cite{PL} & 66.93 & 71.05 & 68.36 & 72.12 & 68.41 & 67.11 & 70.15 & 66.09 & 69.44 & 76.05 & 74.31 & 81.23 & 79.00 & 82.89 & 74.26 & 72.49 & 54.20\\
        
        & TENT~\cite{tent_wang2020} & 67.47 & 70.11 & 68.45 & 72.12 & 68.14 & 69.75 & 72.25 & 68.23 & 70.96 & 82.62 & 78.24 & 80.29 & 69.62 & 76.01 & 87.53 & 73.45 & 54.96\\

        & LAME~\cite{niid_boudiaf2022parameter} & 24.53 & \bf 16.35 & 19.53 & 66.26 & 27.57 & 32.22 & 36.33 & \bf 5.67 & 26.01 & \bf 16.09 & 40.66 & 87.27 & 59.47 & 25.16 & 60.46 & 36.24 & 42.16\\

        & NOTE~\cite{note} & 20.02 & 21.98 & 18.90 & 38.96 & \bf 26.09 & \bf 10.68 & 28.46 & 12.24 & 25.11 & 23.82 & 49.06 & 37.27 & 45.89 & 37.71 & 49.46 & 29.71 & 30.28\\
        
        & TTAC~\cite{su2022revisiting} & 14.79 & 37.62 & 18.28 & 73.19 & 43.39 & 17.16 & 21.72 & 14.83 & 32.26 & 17.29 & 41.73 & 63.49 & 61.48 & 34.94 & 77.21 & 37.96 & 38.07\\

        & RoTTA~\cite{yuan2023robust} & 17.65 & 36.33 & 18.05 & 47.50 & 54.24 & 12.06 & 23.06 & 13.00 & 24.71 & 33.07 & 46.78 & 40.08 & 45.04 & 34.63 & 53.89 & 33.34 & 31.35 \\

        \cmidrule{2-19}
        & \method & \bf 13.09 & 24.98 & \bf 11.75 & \bf 30.07 & 31.05 & 11.44 & \bf 12.24 & 8.89 & \bf 14.57 & 21.54 & \bf 30.74 & \bf 29.22 & \bf 44.68 & \bf 24.53 & \bf 37.62 & \bf 23.10\dtplus{+6.61} & \bf 27.03\dtplus{+3.25}\\
        
    \bottomrule[1.2pt]
    \end{tabular}
    }
    }
\end{table*}

\begin{table*}[t]
    \centering
    \caption{Average classification error on CIFAR100-C while continually adapting to different corruptions at the highest severity 5 with globally and locally class-imbalanced test stream and varying global class distribution (GLI-TTA-V). $I.F.$ is the Imbalanced Factor of Global Class-Imbalance.}
    \label{tab:DLI_VARY_CIFAR100}
    \resizebox{\textwidth}{!}{
    {
    \begin{tabular}{l|l|ccccccccccccccc|cc}
        \toprule[1.2pt]
        \multicolumn{2}{c}{Time} & \multicolumn{15}{l|}{$t\xrightarrow{\hspace*{18.5cm}}$}& \\ \hline
        I.F. & Method & \rotatebox[origin=c]{45}{motion} & \rotatebox[origin=c]{45}{snow} & \rotatebox[origin=c]{45}{fog} & \rotatebox[origin=c]{45}{shot} & \rotatebox[origin=c]{45}{defocus} & \rotatebox[origin=c]{45}{contrast} & \rotatebox[origin=c]{45}{zoom} & \rotatebox[origin=c]{45}{brightness} & \rotatebox[origin=c]{45}{frost} & \rotatebox[origin=c]{45}{elastic} & \rotatebox[origin=c]{45}{glass} & \rotatebox[origin=c]{45}{gaussian} & \rotatebox[origin=c]{45}{pixelate} & \rotatebox[origin=c]{45}{jpeg} & \rotatebox[origin=c]{45}{impulse}
        & Inst.Avg. & Cat.Avg.\\ 
        
        \midrule
        \multirow{9}{*}{10} & TEST & 28.77 & 39.86 & 49.79 & 69.09 & 27.63 & 54.59 & 27.99 & 29.54 & 44.07 & 37.23 & 53.74 & 70.67 & 75.05 & 40.02 & 39.68 & 45.85 & 46.65\\

        & BN~\cite{BN_Stat} & 38.16 & 45.10 & 49.92 & 48.81 & 38.11 & 40.94 & 38.24 & 38.78 & 43.47 & 47.86 & 50.36 & 51.42 & 43.32 & 49.43 & 52.61 & 45.10 & 42.47 \\
        
        & PL~\cite{PL} & 37.72 & 45.05 & 51.70 & 52.17 & 49.07 & 58.69 & 57.28 & 62.25 & 71.39 & 80.31 & 89.22 & 90.71 & 90.82 & 91.31 & 94.69 & 68.16 & 66.62\\
        
        & TENT~\cite{tent_wang2020} & 37.49 & 46.08 & 49.90 & 58.95 & 61.56 & 77.53 & 79.31 & 83.31 & 89.53 & 92.41 & 95.54 & 94.25 & 96.23 & 97.39 & 97.14 & 77.11 & 76.51\\

        & LAME~\cite{niid_boudiaf2022parameter} & \bf 24.41 & 35.55 & 46.44 & 69.22 & \bf 23.43 & 53.74 & \bf 24.10 & 24.95 & 40.87 & \bf 31.09 & 50.85 & 70.49 & 76.78 & \bf 34.96 & 35.37 & 42.82 & 45.35\\

        & NOTE~\cite{note} & 46.39 & 51.68 & 58.85 & 60.81 & 49.10 & 44.32 & 46.65 & 50.77 & 55.80 & 59.31 & 66.49 & 68.58 & 62.77 & 68.34 & 81.11 & 58.07 & 58.46\\

        & TTAC~\cite{su2022revisiting} & 27.40 & 40.97 & 42.80 & 52.53 & 31.06 & 43.14 & 30.37 & 27.22 & 35.99 & 37.00 & 41.36 & 46.39 & 48.48 & 39.78 & \bf 33.90 & 38.56 & 38.68\\

        & RoTTA~\cite{yuan2023robust} & 33.49 & 42.91 & 46.70 & 48.84 & 34.29 & 42.39 & 35.96 & 27.79 & 37.72 & 39.76 & 42.11 & 42.03 & 37.77 & 38.54 & 43.06 & 39.56 & 39.77\\
        
        \cmidrule{2-19}
        & \method & 26.57 & \bf 33.88 & \bf 38.47 & \bf 37.85 & 29.18 & \bf 29.95 & 26.37 & \bf 24.56 & \bf 30.29 & 34.70 & \bf 39.22 & \bf 39.86 & \bf 34.31 & 37.18 & 40.38 & \bf 33.52\dtplus{+5.04} & \bf 34.49\dtplus{+4.19}\\

        \midrule
        \multirow{9}{*}{100} & TEST & 26.75 & 39.04 & 49.74 & 68.37 & 26.61 & 54.67 & 28.02 & 28.30 & 43.17 & 37.31 & 53.97 & 67.81 & 76.96 & 38.39 & 40.97 & 45.34 & 46.94\\

        & BN~\cite{BN_Stat} & 37.59 & 45.00 & 51.81 & 46.97 & 38.39 & 41.58 & 38.85 & 35.85 & 43.22 & 47.40 & 52.84 & 49.88 & 43.64 & 49.74 & 54.48 & 45.15 & 38.80\\

        & PL~\cite{PL} & 35.52 & 44.11 & 48.76 & 47.82 & 43.22 & 46.55 & 47.02 & 45.75 & 51.01 & 57.06 & 61.47 & 60.44 & 60.39 & 69.03 & 74.24 & 52.83 & 48.39\\
        
        & TENT~\cite{tent_wang2020} & 34.44 & 44.30 & 49.41 & 48.43 & 49.74 & 58.09 & 55.75 & 60.86 & 69.78 & 77.24 & 81.14 & 81.65 & 86.06 & 91.37 & 93.01 & 65.42 & 63.48\\

        & LAME~\cite{niid_boudiaf2022parameter} & \bf 22.62 & 34.30 & 46.18 & 69.26 & \bf 21.68 & 53.59 & \bf 24.73 & 23.37 & 40.83 & \bf 31.58 & 51.10 & 67.53 & 79.26 & \bf 34.07 & \bf 36.98 & 42.47 & 47.82\\

        & NOTE~\cite{note} & 45.52 & 49.55 & 58.75 & 56.41 & 46.55 & 43.69 & 44.96 & 47.25 & 51.43 & 57.02 & 64.62 & 67.06 & 61.33 & 60.21 & 73.02 & 55.16 & 55.95\\
        
        & TTAC~\cite{su2022revisiting} & 27.83 & 48.10 & 46.13 & 55.94 & 26.89 & 46.97 & 29.61 & 26.84 & 38.90 & 39.00 & 43.78 & 48.33 & 61.80 & 49.60 & 41.30 & 42.07 & 41.05\\

        & RoTTA~\cite{yuan2023robust} & 30.41 & 41.67 & 48.80 & 49.32 & 37.21 & 38.85 & 42.56 & 25.06 & 43.50 & 50.26 & 44.77 & 46.27 & 48.90 & 39.18 & 46.22 & 42.20 & 39.93\\

        \cmidrule{2-19}
        & \method & 24.78 & \bf 32.94 & \bf 38.48 & \bf 36.37 & 30.46 & \bf 30.78 & 28.77 & \bf 22.15 & \bf 30.92 & 36.65 & \bf 41.44 & \bf 39.00 & \bf 37.35 & 35.85 & 40.50 & \bf 33.76\dtplus{+8.31} & \bf 34.63\dtplus{+4.17}\\

        \midrule
        \multirow{9}{*}{200} & TEST & 25.83 & 39.37 & 49.41 & 68.26 & 26.85 & 54.35 & 27.87 & 28.14 & 43.23 & 37.65 & 53.81 & 66.43 & 77.71 & 37.59 & 40.92 & 45.16 & 40.61\\

        & BN~\cite{BN_Stat} & 36.68 & 46.62 & 51.18 & 46.94 & 41.78 & 40.76 & 39.42 & 35.77 & 42.80 & 47.26 & 52.04 & 51.40 & 43.72 & 49.73 & 54.46 & 45.37 & 33.45\\

        & PL~\cite{PL} & 35.28 & 43.45 & 49.46 & 45.54 & 44.90 & 48.98 & 47.74 & 44.36 & 51.88 & 56.93 & 66.22 & 64.45 & 61.55 & 67.88 & 76.53 & 53.68 & 44.28\\
        
        & TENT~\cite{tent_wang2020} &  34.32 & 44.79 & 50.21 & 47.91 & 49.14 & 55.32 & 52.31 & 54.78 & 61.65 & 72.40 & 79.00 & 74.76 & 80.77 & 88.29 & 91.08 & 62.45 & 53.57\\

        & LAME~\cite{niid_boudiaf2022parameter} & \bf 21.97 & 34.75 & 46.51 & 69.33 & \bf 22.13 & 53.65 & \bf 24.01 & 23.20 & 41.14 & \bf 31.79 & 49.68 & 66.43 & 79.70 & \bf 32.12 & \bf 37.00 & 42.23 & 41.45\\

        & NOTE~\cite{note} & 45.33 & 49.19 & 59.08 & 57.25 & 45.97 & 43.29 & 45.76 & 46.08 & 50.54 & 55.75 & 62.78 & 64.61 & 60.10 & 59.13 & 71.64 & 54.43 & 48.65\\

        & TTAC~\cite{su2022revisiting} & 26.96 & 49.19 & 48.44 & 55.59 & 27.39 & 47.10 & 29.54 & 27.18 & 39.90 & 40.28 & 43.18 & 49.68 & 63.16 & 51.50 & 43.98 & 42.87 & 35.80\\

        & RoTTA~\cite{yuan2023robust} & 29.27 & 42.80 & 52.95 & 47.85 & 39.26 & 39.74 & 47.53 & 25.35 & 43.50 & 53.54 & 45.86 & 47.21 & 51.99 & 40.82 & 45.86 & 43.57 & 35.82 \\

        \cmidrule{2-19}
        & \method & 23.15 & \bf 33.83 & \bf 40.44 & \bf 37.27 & 31.53 & \bf 31.53 & 28.84 & \bf 22.50 & \bf 30.72 & 38.83 & \bf 40.60 & \bf 38.88 & \bf 39.96 & 35.23 & 41.08 & \bf 34.29\dtplus{+7.94} & \bf 30.22\dtplus{+3.23}\\
        
    \bottomrule[1.2pt]
    \end{tabular}
    }
    }
\end{table*}

\begin{table*}[t]
    \centering
    \caption{Average classification error of the task CIFAR10 $\to$ CIFAR10-C while continually adapting to different corruptions at the highest severity 5 with correlatively sampled test stream under the PTTA setup.}
    \label{tab:CIFAR10}
    \resizebox{\textwidth}{!}{
    {
    \begin{tabular}{l|ccccccccccccccc|c}
        \toprule[1.2pt]
        Time & \multicolumn{15}{l|}{$t\xrightarrow{\hspace*{18.5cm}}$}& \\ \hline
        Method & \rotatebox[origin=c]{45}{motion} & \rotatebox[origin=c]{45}{snow} & \rotatebox[origin=c]{45}{fog} & \rotatebox[origin=c]{45}{shot} & \rotatebox[origin=c]{45}{defocus} & \rotatebox[origin=c]{45}{contrast} & \rotatebox[origin=c]{45}{zoom} & \rotatebox[origin=c]{45}{brightness} & \rotatebox[origin=c]{45}{frost} & \rotatebox[origin=c]{45}{elastic} & \rotatebox[origin=c]{45}{glass} & \rotatebox[origin=c]{45}{gaussian} & \rotatebox[origin=c]{45}{pixelate} & \rotatebox[origin=c]{45}{jpeg} & \rotatebox[origin=c]{45}{impulse}
        & Avg. \\ 
        
        \midrule
        TEST & 34.8 & 25.1 & 26.0 & 65.7 & 46.9 & 46.7 & 42.0 & 9.3 & 41.3 & 26.6 & 54.3 & 72.3 & 58.5 & 30.3 & 72.9 & 43.5\\

        BN~\cite{BN_Stat} & 73.2 & 73.4 & 72.7 & 77.2 & 73.7 & 72.5 & 72.9 & 71.0 & 74.1 & 77.7 & 80.0 & 76.9 & 75.5 & 78.3 & 79.0 & 75.2 \\

        PL~\cite{PL} & 73.9 & 75.0 & 75.6 & 81.0 & 79.9 & 80.6 & 82.0 & 83.2 & 85.3 & 87.3 & 88.3 & 87.5 & 87.5 & 87.5 & 88.2 & 82.9 \\

        TENT~\cite{tent_wang2020} & 74.3 & 77.4 & 80.1 & 86.2 & 86.7 & 87.3 & 87.9 & 87.4 & 88.2 & 89.0 & 89.2 & 89.0 & 88.3 & 89.7 & 89.2 & 86.0 \\
        
        LAME~\cite{niid_boudiaf2022parameter} & 29.5 &  19.0 & {20.3} & 65.3 & 42.4 & 43.4 & 36.8 &  \bf 5.4 & 37.2 &  18.6 & 51.2 & 73.2 & 57.0 &  22.6 & 71.3 & 39.5 \\

        CoTTA~\cite{cotta} & 77.1 & 80.6 & 83.1 & 84.4 & 83.9 & 84.2 & 83.1 & 82.6 & 84.4 & 84.2 & 84.5 & 84.6 & 82.7 & 83.8 & 84.9 & 83.2 \\

        NOTE~\cite{note} &  18.0 & 22.1 & 20.6 & 35.6 & 26.9 &  13.6 & 26.5 & 17.3 & 27.2 & 37.0 & 48.3 & 38.8 & 42.6 & 41.9 & 49.7 & 31.1 \\
        
        TTAC~\cite{su2022revisiting} & 20.7 & 18.7 & 13.1 & 38.9 & 16.9 & 17.4 & 12.7 & 8.6 & 19.5 & 24.6 & 32.1 & 33.9 & 25.0 & 24.5 & 38.8 & 23.0\\
        
        RoTTA~\cite{yuan2023robust} & 18.1 & 21.3 &  18.8 &  33.6 &  23.6 & 16.5 &  15.1 & 11.2 &  21.9 & 30.7 &  39.6 &  26.8 &  33.7 & 27.8 &  39.5 & 25.2 \\

        \midrule
        \method & \bf 14.4 & \bf 13.3 & \bf 9.3 & \bf 20.7 & \bf 12.8 & \bf 10.5 & \bf 10.3 & 6.8 & \bf 13.1 & \bf 18.1 & \bf 24.8 & \bf 20.4 & \bf 18.2 & \bf 21.3 & \bf 28.4 & \bf 16.1\dtplus{+6.9}\\
        
    \bottomrule[1.2pt]
    \end{tabular}
    }
    }
\end{table*}

\begin{table*}[t]
    \centering
    \caption{Average classification error of the task CIFAR100 $\to$ CIFAR100-C while continually adapting to different corruptions at the highest severity 5 with correlatively sampled test stream under the PTTA setup.}
    \label{tab:CIFAR100}
    \resizebox{\textwidth}{!}{
    {
    \begin{tabular}{l|ccccccccccccccc|c}
        \toprule[1.2pt]
        Time & \multicolumn{15}{l|}{$t\xrightarrow{\hspace*{18.5cm}}$}& \\ \hline
        Method & \rotatebox[origin=c]{45}{motion} & \rotatebox[origin=c]{45}{snow} & \rotatebox[origin=c]{45}{fog} & \rotatebox[origin=c]{45}{shot} & \rotatebox[origin=c]{45}{defocus} & \rotatebox[origin=c]{45}{contrast} & \rotatebox[origin=c]{45}{zoom} & \rotatebox[origin=c]{45}{brightness} & \rotatebox[origin=c]{45}{frost} & \rotatebox[origin=c]{45}{elastic} & \rotatebox[origin=c]{45}{glass} & \rotatebox[origin=c]{45}{gaussian} & \rotatebox[origin=c]{45}{pixelate} & \rotatebox[origin=c]{45}{jpeg} & \rotatebox[origin=c]{45}{impulse}
        & Avg. \\ 
        
        \midrule
        TEST & 30.8 & 39.5 & 50.3 & 68.0 & {29.3} & 55.1 & 28.8 & 29.5 & 45.8 & 37.2 & 54.1 & 73.0 & 74.7 & 41.2 & {39.4} & 46.4 \\

        BN~\cite{BN_Stat} & 48.5 & 54.0 & 58.9 & 56.2 & 46.4 & {48.0} & 47.0 & 45.4 & 52.9 & 53.4 & 57.1 & 58.2 & 51.7 & 57.1 & 58.8 & 52.9 \\

        PL~\cite{PL} & 50.6 & 62.1 & 73.9 & 87.8 & 90.8 & 96.0 & 94.8 & 96.4 & 97.4 & 97.2 & 97.4 & 97.4 & 97.3 & 97.4 & 97.4 & 88.9 \\

        TENT~\cite{tent_wang2020} & 53.3 & 77.6 & 93.0 & 96.5 & 96.7 & 97.5 & 97.1 & 97.5 & 97.3 & 97.2 & 97.1 & 97.7 & 97.6 & 98.0 & 98.3 & 92.8 \\

        LAME~\cite{niid_boudiaf2022parameter} & \bf 22.4 & \bf 30.4 & {43.9} & 66.3 &  \bf 21.3 & 51.7 & \bf 20.6 & \bf 21.8 & {39.6} & 28.0 & {48.7} & 72.8 & 74.6 & \bf 33.1 & 32.3 & {40.5} \\

        CoTTA~\cite{cotta} & 49.2 & 52.7 & 56.8 & {53.0} & 48.7 & 51.7 & 49.4 & 48.7 & 52.5 & 52.2 & 54.3 & {54.9} & {49.6} & 53.4 & 56.2 & 52.2 \\

        NOTE~\cite{note} & 45.7 & 53.0 & 58.2 & 65.6 & 54.2 & 52.0 & 59.8 & 63.5 & 74.8 & 91.8 & 98.1 & 98.3 & 96.8 & 97.0 & 98.2 & 73.8  \\
    
        TTAC~\cite{su2022revisiting} & 28.0 & 35.4 & 39.0 & 45.5 & 27.3 & 34.0 & 27.1 & 24.9 & \bf 32.4 & 32.9 & 39.2 & 43.7 & 34.3 & 37.0 & \bf 31.3 & 34.1\\

        ROTTA~\cite{yuan2023robust} & {31.8} & {36.7} & 40.9 & 42.1 & 30.0 & 33.6 & {27.9} & {25.4} & 32.3 & {34.0} & 38.8 & \bf 38.7 & \bf 31.3 & {38.0} & 42.9 & 35.0 \\

        \midrule
        \method & 28.3 & 32.6 & \bf 37.7 & \bf 38.4 & 28.6 & \bf 28.3 & 26.2 & 25.1 & \bf 31.4 & 34.0 & \bf 38.7 & 39.4 & 32.1 & 38.4 & 39.5 & \bf 33.3\dtplus{+0.8}\\
        
    \bottomrule[1.2pt]
    \end{tabular}
    }
    }
\end{table*}

\end{document}